%% file: acl_latex.tex
\documentclass[11pt]{article}

\PassOptionsToPackage{hyperfootnotes=false}{hyperref}
\usepackage[final]{acl}

\usepackage{times}
\usepackage{latexsym}

\usepackage[T1]{fontenc}

\usepackage[utf8]{inputenc}

\usepackage{microtype}

\usepackage{inconsolata}

\usepackage{graphicx}

\usepackage{amsmath}
\DeclareMathOperator*{\argmax}{arg\,max}

\usepackage{amssymb}
\usepackage[table]{xcolor}
\usepackage{booktabs}
\usepackage{multirow}
\usepackage{subcaption}
\usepackage[most]{tcolorbox}

\newcommand{\squishlist}{
   \begin{list}{\small$\bullet$}
    { \setlength{\itemsep}{0pt}
      \setlength{\parsep}{3pt}
      \setlength{\topsep}{3pt}
      \setlength{\partopsep}{0pt}
      \setlength{\leftmargin}{1.5em}
      \setlength{\labelwidth}{1em}
      \setlength{\labelsep}{0.5em} } }

\newcounter{Lcount}
\newcommand{\squishlisttwo}{
   \begin{list}{\arabic{Lcount}. }
    { \usecounter{Lcount}
      \setlength{\itemsep}{0pt}
      \setlength{\parsep}{0pt}
      \setlength{\topsep}{0pt}
      \setlength{\partopsep}{0pt}
      \setlength{\leftmargin}{2em}
      \setlength{\labelwidth}{1.5em}
      \setlength{\labelsep}{0.5em} } }

\newcommand{\squishend}{
    \end{list}  }

\newcommand{\framework}{\textsc{FoCUS}}
\newcommand{\benchmark}{\textsc{SCoPE}}

\title{Controllable Image Captioning with Prompt-Conditioned Scene Rewards}

\newcommand{\sparagraph}[1]{\vspace{3pt}\par\noindent\textbf{#1}}

\author{
  \textbf{Jongyeop Hyun}\textsuperscript{1}\thanks{Equal contribution.}
  \quad
  \textbf{Taeyoung Kim}\textsuperscript{1}\footnotemark[1]
  \quad
  \textbf{Hyounghun Kim}\textsuperscript{1,2}
  \\
  \textsuperscript{1}Graduate School of Artificial Intelligence, POSTECH
  \\
  \textsuperscript{2}Department of Computer Science and Engineering, POSTECH
  \\
  \texttt{\{mldljyh,kty4119,h.kim\}@postech.ac.kr}
}

\begin{document}
\maketitle

\input{latex/sec/0_Abstract}
\input{latex/sec/1_Intro}
\input{latex/sec/figures/focus_overview}
\input{latex/sec/2_Related_Work}
\input{latex/sec/4_Method}
\input{latex/sec/5_Benchmark}
\input{latex/sec/6_Experiments}
\input{latex/sec/7_Result}
\input{latex/sec/8_Conclusion}

\input{latex/sec/98_Limitations}
\input{latex/sec/99_Ethics_Statement}

\section*{Acknowledgments}
This work was supported by the Institute of Information \& Communications Technology Planning \& Evaluation (IITP) grant funded by the Korea government (MSIT) (No.~RS-2019-II191906, Artificial Intelligence Graduate School Program (POSTECH)) and by the Ministry of Education of the Republic of Korea and the National Research Foundation of Korea (NRF-2022S1A5A2A03052246). We thank the reviewers and the Action Editor for their valuable feedback.

\bibliography{custom}

\clearpage
\appendix
\input{latex/sec/A_CompreCap}
\input{latex/sec/A_Object_Matching}
\input{latex/sec/A_SCOPE_Construction}
\input{latex/sec/A_SCoPE-Stats}
\input{latex/sec/A_overall_interpre}
\input{latex/sec/A_Human_Alignment}
\input{latex/sec/A_Exp_Details}
\input{latex/sec/A_more_ablations}
\input{latex/sec/A_Commercial}
\input{latex/sec/A_Prompt}
\input{latex/sec/A_Examples}
\end{document}

%% file: latex/sec/0_Abstract.tex
\begin{abstract}
Large Vision-Language Models produce fluent image descriptions but offer limited semantic control: users cannot reliably specify whether captions should emphasize attributes, relations, or particular image regions. We present \textbf{F}ine-grained Capti\textbf{o}ning \textbf{C}ontrol \textbf{U}sing \textbf{S}cene Rewards (\textbf{\framework{}}), a controllable image captioning method that lets users steer captions toward specific semantic emphases through natural-language control prompts. The core idea is a prompt-conditioned control objective based on scene-graph-aligned component scores. Generated captions are parsed and aligned to scene-graph components---objects, attributes, and relations---and these components are differentially weighted, including negative weights, according to the requested emphasis. We optimize this objective with GRPO and further improve its reliability through a stricter object validity threshold and reasoning-based verification for attribute and relation scoring. To evaluate controllability, we introduce \textbf{S}emantic \textbf{Co}ntrol and \textbf{P}recision \textbf{E}valuation (\textbf{\benchmark{}}), a benchmark with contrastive Include/Avoid constraints for measuring both target content coverage and out-of-scope suppression. Experiments on two VLM backbones show that \framework{} consistently improves controllability and fine-grained caption quality while largely maintaining general caption performance.\footnote{The code and dataset are publicly available at \url{https://focus-emnlp2026.github.io/}.}

\end{abstract}

%% file: latex/sec/1_Intro.tex
\section{Introduction}
\label{sec:introduction}

\input{latex/sec/figures/teaser}

Image captioning has advanced with the emergence of Large Vision-Language Models (LVLMs) that couple strong visual encoders with powerful large language models, enabling fluent long-form descriptions and instruction-following behavior~\citep{li2023blip2,liu2023visual,bai2023qwenvl,chen2024intern}. In parallel, richer supervision---such as detailed caption datasets and scene-level annotations---has improved coverage and specificity~\citep{chen2024sharegpt4v,onoe2024docci,pont-tuset2020connecting}. Despite this progress, a practical limitation remains: users often need \emph{semantic control} over what a caption emphasizes. For the same image, one may want an attribute-centric description (colors, materials), a relation-centric description (spatial/functional relationships), or a foreground/background-focused description. Current LVLM captioners produce a single ``best overall'' caption and offer limited mechanisms to steer \emph{what to focus on}. Figure~\ref{fig:teaser} shows that zero-shot prompting often defaults to a generic scene description rather than the requested semantic focus.

Existing controllable captioning methods usually require structured controls at inference time, such as length tokens~\citep{deng2020lengthcontrollable}, regions or boxes~\citep{cornia2019show}, or formal semantic graphs~\citep{basioti2024cicbartssa}. Natural-language prompting is a more natural interface, but prompting alone is often unreliable for fine-grained emphasis and suppression, with models drifting toward generic high-probability content.

A natural alternative is to decompose captions into interpretable semantic units and let the prompt determine which units to reward or penalize. Scene graphs---objects, attributes, and relations---provide such a decomposition, and recent work has shown their value for fine-grained caption evaluation~\citep{dong2024benchmarking,lu2025benchmarking}. However, they have mainly been used for post-hoc evaluation or prompt-agnostic training, not as a prompt-conditioned control signal that rewards requested content while suppressing off-scope content.

We propose \textbf{F}ine-grained Capti\textbf{o}ning \textbf{C}ontrol \textbf{U}sing \textbf{S}cene Rewards (\textbf{\framework{}}), a controllable captioning method built around a prompt-conditioned scene-graph objective. Given a natural-language prompt specifying a semantic emphasis (e.g., attribute-, relation-, foreground-, background-focused, or general), \framework{} parses a generated caption into scene-graph components and aggregates scores using prompt-specific signed weights: positive weights reward requested content, while negative weights penalize off-scope content. This turns scene-graph decomposition from an evaluation tool into an interpretable learning signal for caption generation, without architectural changes or structured side inputs. In our implementation, we optimize this objective with GRPO~\citep{shao-etal-2024-deepseekmath}.
To reduce reward noise, we use stricter object-validity checks and reasoning-based attribute/relation verification.

Evaluation is another challenge. Standard captioning metrics measure overall quality, but not \emph{contrastive controllability}: covering requested content while avoiding off-scope content~\citep{anderson2016spice}. Existing fine-grained evaluators provide component-level feedback, but they do not explicitly test this include-vs.-avoid behavior~\citep{dong2024benchmarking,lu2025benchmarking}. We therefore introduce \textbf{S}emantic \textbf{Co}ntrol and \textbf{P}recision \textbf{E}valuation (\textbf{\benchmark{}}), a benchmark with category-specific \emph{Include} and \emph{Avoid} lists derived from curated captions. \benchmark{} measures target coverage, off-scope suppression, and factual consistency.

Our main contributions are:
\squishlist
    \item We propose \framework{}, a controllable image captioning method based on a prompt-conditioned control objective over scene-graph components, with signed weights that both emphasize requested content and suppress off-scope content.
    \item We improve the reliability of scene-graph rewards for learning through stricter object matching and reasoning-based verification for attributes and relations, and show gains over holistic and fixed-weight reward baselines.
    \item We introduce \benchmark{}, a benchmark for controllable captioning with contrastive Include/Avoid constraints for measuring coverage, adherence, and contradiction-based faithfulness.
\squishend

%% file: latex/sec/figures/teaser.tex
\begin{figure}[!t]
  \vspace{-2mm}
  \centering
  \includegraphics[width=0.98\columnwidth]
  {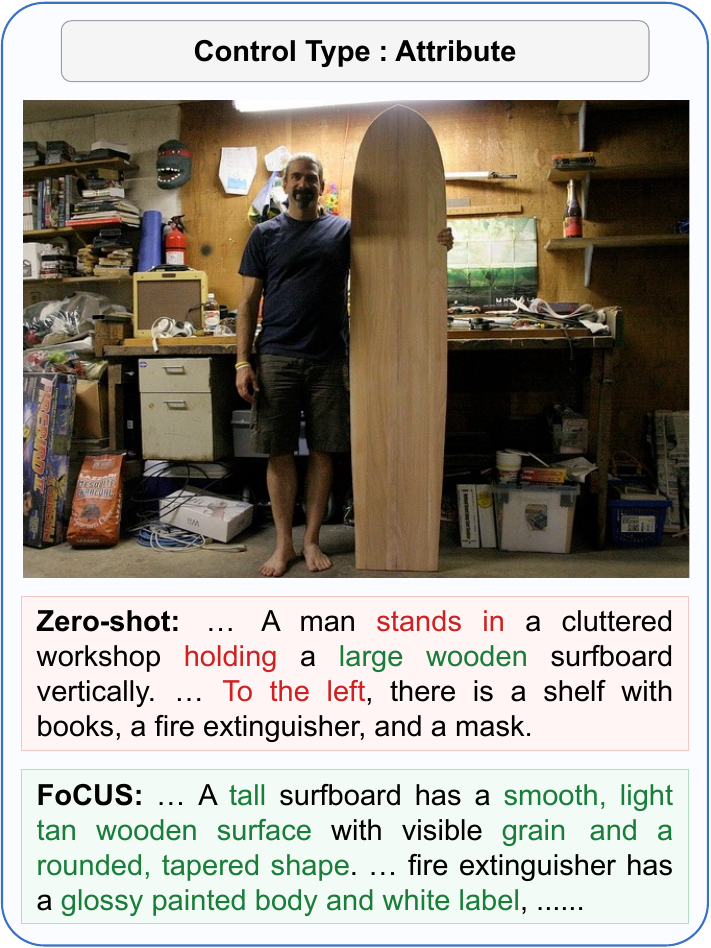}
  \caption{Comparison of Zero-shot and \framework{} generation. Zero-shot defaults to a generic scene description (red), whereas \framework{} emphasizes object attributes (green) when generating an Attribute-focused caption.}
  \label{fig:teaser}
  \vspace{-4mm}
\end{figure}

%% file: latex/sec/figures/focus_overview.tex
\begin{figure*}[!t]
\vspace{-2mm}
  \centering
  \includegraphics[width=\textwidth]{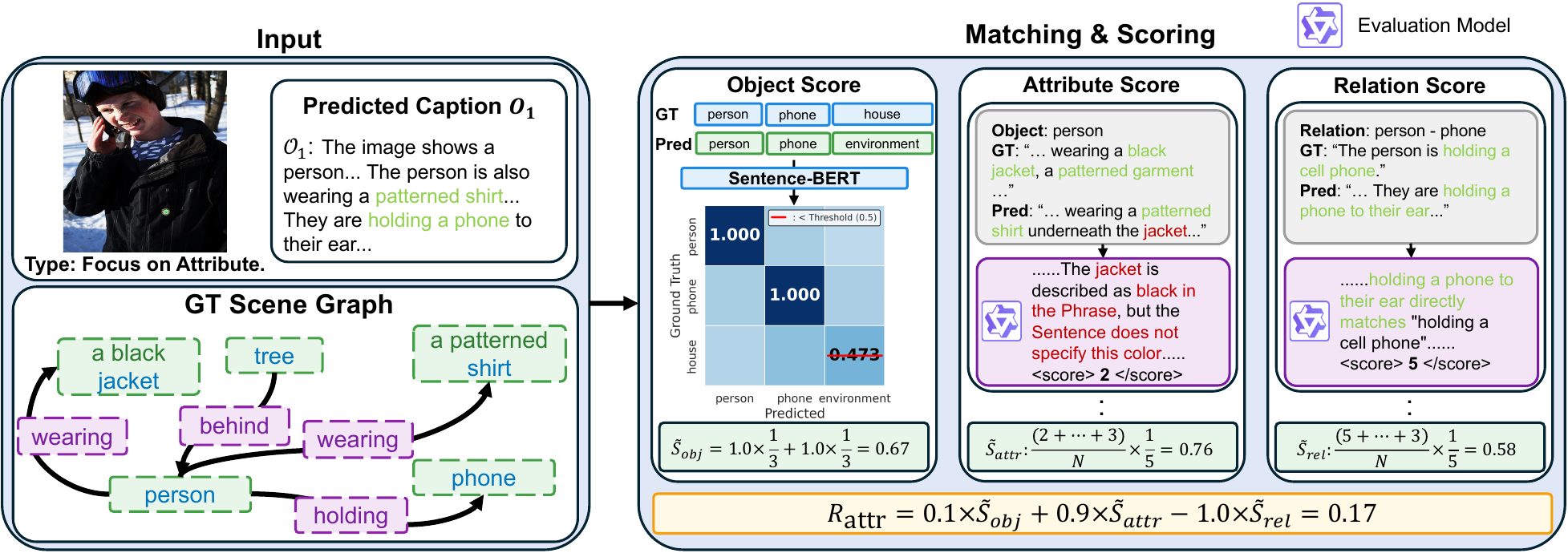}
  \caption{Overview of the \framework{} pipeline. During training, given an image and a natural-language control prompt, \framework{} aligns the generated caption with scene-graph annotations and computes component scores for objects, attributes, and relations. These scores are aggregated with prompt-conditioned signed weights to form the control objective, which emphasizes requested content and suppresses off-scope content.}
  \label{fig:method}
  \vspace{-2mm}
\end{figure*}

%% file: latex/sec/2_Related_Work.tex
\section{Related Work}
\label{sec:related_work}

\sparagraph{Large Vision-Language Models and Detailed Captioning.}
Modern LVLMs couple strong visual encoders with large language models and produce fluent, detailed captions~\citep{li2023blip2,liu2023visual,bai2023qwenvl,chen2024intern}. This progress is supported by richer supervision sources, including ShareGPT4V~\citep{chen2024sharegpt4v}, DOCCI~\citep{onoe2024docci}, and Localized Narratives~\citep{pont-tuset2020connecting}. However, these models typically default to a single ``best overall'' description and offer limited, unreliable control over which semantics (e.g., attributes vs.\ relations, foreground vs.\ background) are emphasized.

\sparagraph{Controllable Image Captioning.}
Prior controllable captioning methods usually rely on structured side inputs at inference time, such as explicit length signals~\citep{deng2020lengthcontrollable}, region-level inputs (e.g., boxes or masks)~\citep{cornia2019show}, or formal semantic specifications such as AMR graphs~\citep{basioti2024cicbartssa}. Our setting instead uses natural-language control prompts and learns to realize them through a prompt-conditioned objective over scene-graph components, without additional structured inputs at inference time or architectural changes.

\sparagraph{Training Objectives for Caption Alignment.}
Caption generation has been optimized beyond standard cross-entropy using metric-based objectives such as SCST with CIDEr~\citep{rennie2017selfcritical,vedantam2015cidera}, embedding-based signals such as CLIPScore~\citep{hessel2021clipscore}, and preference-based objectives such as DPO~\citep{rafailov2023direct}. Scene-graph-based metrics such as SPICE, CAPTURE, and CompreCap provide finer object-, attribute-, and relation-level feedback~\citep{anderson2016spice,dong2024benchmarking,lu2025benchmarking}, and recent work such as SC-Captioner~\citep{zhang2025sccaptioner} uses scene-graph components as compositional supervision. Our work extends this line by making scene-graph rewards prompt-conditioned: object, attribute, and relation scores are combined with prompt-specific signed weights to produce captions with user-specified semantic emphases.

\sparagraph{Captioning Benchmarks and Fine-grained Evaluation.}
Captioning benchmarks commonly measure overall quality with metrics such as CIDEr~\citep{vedantam2015cidera} and SPICE~\citep{anderson2016spice}. More recent evaluations, including CAPTURE~\citep{dong2024benchmarking} and CompreCap~\citep{lu2025benchmarking}, provide component-wise analysis but do not explicitly test \emph{contrastive controllability}---maximizing requested content while suppressing off-scope content. We address this gap with \benchmark{}, which introduces category-specific \textit{Include} and \textit{Avoid} constraints to measure both coverage (recall) and suppression (precision) under user-specified semantic emphases.

%% file: latex/sec/4_Method.tex
\section{Method}
\label{sec:method}

We present \framework{}, a controllable image captioning method built around a prompt-conditioned, scene-graph-aligned objective. Given a natural-language control prompt, \framework{} decomposes a generated caption into objects, attributes, and relations, then aggregates their scores with prompt-specific signed weights to reward requested content and penalize off-scope content. We optimize this objective with a two-stage SFT+GRPO~\citep{shao-etal-2024-deepseekmath} pipeline. Figure~\ref{fig:method} illustrates \framework{}.

\subsection{Scene-Graph-Aligned Component Scores}

Our scoring function is inspired by CompreCap~\citep{lu2025benchmarking}, which decomposes captions into objects, attributes, and relations for component-wise scoring, but we adapt it for reward-based training. Directly optimizing such a pipeline in RL can be noisy: embedding-based object matching may drift without a validity threshold, and zero-shot LLM judges may over-reward generic descriptions. We therefore introduce stricter object validity checks and stronger Chain-of-Thought-based verification~\citep{wei2022chainofthought} for attribute and relation scoring with modern LLM judges. Additional background and motivation are in Appendix~\ref{app:comprecap}.

We parse the generated caption into scene-graph elements using spaCy~\citep{honnibal2020spacy} for noun extraction and lemmatization, then compute scores at three semantic levels.

\sparagraph{Object Score ($S_\text{obj}$).}
We compute the soft coverage of annotated objects mentioned in the generated caption. Let $\mathcal{C}=\{c_i\}_{i=1}^{n}$ be the candidate object nouns extracted from the generated caption, and let $\mathcal{G}=\{g_j\}_{j=1}^{m}$ be the ground-truth object categories from the annotated scene-graph. Using Sentence-BERT~\citep{reimers2019sentencebert}, we compute pairwise similarities and construct a mutual-best match indicator matrix $\mathbf{S}'\in\{0,1\}^{n\times m}$ with a validity threshold $\tau=0.5$ (details in Appendix~\ref{app:object_matching}). The object score is defined as the fraction of ground-truth objects successfully matched:
\begin{equation}
S_\text{obj} = \frac{1}{m}\sum_{j=1}^{m}\max_{i}\, S'_{i,j}.
\label{eq:object_score}
\end{equation}

\sparagraph{Attribute Score ($S_\text{attr}$).}
For objects matched at the object level, we evaluate the quality of attribute descriptions. Let $\mathcal{J}$ denote the index set of matched ground-truth objects. For each $j\in\mathcal{J}$, we employ \texttt{Qwen3-30B-A3B-Instruct}~\citep{yang2025qwen3} with CoT prompting to produce an integer attribute alignment score $a_j\in\{0,1,\ldots,5\}$. The attribute score is the mean over all matched objects:
\begin{equation}
S_\text{attr} = \frac{1}{|\mathcal{J}|}\sum_{j\in\mathcal{J}} a_j.
\label{eq:attribute_score}
\end{equation}
If no objects are matched, we set $S_\text{attr}=0$.

\input{latex/sec/figures/scope_overview}

\sparagraph{Relation Score ($S_\text{rel}$).}
For relation evaluation, we gather subcaptions containing all objects involved in each annotated directed relation and employ the same CoT-based judge to assess whether the generated caption correctly captures the directional relationship, producing a score $r_k\in\{0,1,\ldots,5\}$ for each relation $k$. The relation score is:
\begin{equation}
S_\text{rel}=\frac{1}{|\mathcal{E}|}\sum_{k\in\mathcal{E}} r_k,
\label{eq:relation_score}
\end{equation}
where $\mathcal{E}$ denotes the set of annotated directed relations. If $|\mathcal{E}|=0$, we set $S_\text{rel}=0$. For stable optimization, we linearly scale all component scores to $[0,1]$ (denoted as $\tilde{S}$), specifically setting $\tilde{S}_\text{obj}=S_\text{obj}$, $\tilde{S}_\text{attr}=S_\text{attr}/5$, and $\tilde{S}_\text{rel}=S_\text{rel}/5$. The exact judge prompts used for attribute and relation scoring are provided in Appendix~\ref{app:training_reward_prompts}.

\sparagraph{Foreground and Background Scores.}
To enable spatially aware controllable captioning, we classify objects as foreground (salient, central) or background (contextual, peripheral). We compute $\tilde{S}_\text{fg}$ and $\tilde{S}_\text{bg}$ by aggregating object, attribute, and relation scores over the corresponding subset, using the same weights as the general caption reward:
\begin{equation}
\tilde{S}_\text{fg/bg} = 0.25 \tilde{S}_\text{obj}^{fg/bg} + 0.35 \tilde{S}_\text{attr}^{fg/bg} + 0.40 \tilde{S}_\text{rel}^{fg/bg}.
\end{equation}

\subsection{Prompt-Conditioned Control Objective}
The central mechanism in \framework{} is a prompt-conditioned aggregation of scene-graph component scores. Given a generated caption $y$, a control prompt category $p$, and scene-graph annotations $z^*$ for the image, we define
\begin{equation}
R(y \mid p, z^*) = \sum_{k \in \mathcal{K}} w_k(p)\,\tilde{S}_k(y; z^*),
\end{equation}
where $\mathcal{K}$ denotes the relevant semantic components, $\tilde{S}_k(y; z^*)\in[0,1]$ is the normalized score for component $k$, and $w_k(p)$ is a prompt-specific weight. Below, we omit the dependence of $\tilde{S}_k$ on $(y, z^*)$ for brevity.
Signed weights let the objective reward requested content, suppress off-scope content, and adapt its component emphasis to the user's request, unlike fixed or prompt-agnostic rewards.

We consider five prompt categories: \emph{General}, \emph{Attribute-}, \emph{Relation-}, \emph{Foreground-}, and \emph{Background-Focused} (see Appendix~\ref{app:prompt_categories} for descriptions). We use CompreCap's unified weighting scheme for the general prompt, which assigns larger weights to more challenging semantic components and was designed to align better with human judgments. The remaining prompts use signed weights to induce category-specific emphasis and suppression:
\begin{align}
\begin{alignedat}{7}
R_\text{general}
  &= {}& 0.25\,\tilde{S}_\text{obj}
  &+{}& 0.35\,\tilde{S}_\text{attr}
  &+{}& 0.40\,\tilde{S}_\text{rel}, \\
R_\text{attr}
  &= {}& 0.10\,\tilde{S}_\text{obj}
  &+{}& 0.90\,\tilde{S}_\text{attr}
  &-{}& 1.00\,\tilde{S}_\text{rel}, \\
R_\text{rel}
  &= {}& 0.10\,\tilde{S}_\text{obj}
  &-{}& 1.00\,\tilde{S}_\text{attr}
  &+{}& 0.90\,\tilde{S}_\text{rel},
\end{alignedat}
\\[0.3em]
\begin{alignedat}{5}
R_\text{fg}
  &= {}& \tilde{S}_\text{fg}
  &-{}& \tilde{S}_\text{bg}, \qquad
R_\text{bg}
  &= {}& \tilde{S}_\text{bg}
  &-{}& \tilde{S}_\text{fg}.
\end{alignedat}
\end{align}

\subsection{Optimization with GRPO}
We optimize the prompt-conditioned control objective above with GRPO. For an image--prompt--annotation triple $(x,p,z^*)$ and generated caption $y \sim \pi_\theta(\cdot \mid x,p)$, we optimize
\begin{equation}
\begin{aligned}
\max_\theta\;&
\mathbb{E}_{\substack{
(x,p,z^*)\sim\mathcal{D}\\
y\sim\pi_\theta(\cdot|x,p)
}}
\Big[
R(y \mid p, z^*) \\
&\quad
- \beta \mathrm{D}_\mathrm{KL}\big(
\pi_\theta(\cdot|x,p)
\,\|\, 
\pi_\mathrm{ref}(\cdot|x,p)
\big)
\Big],
\end{aligned}
\end{equation}
where $R(y \mid p, z^*)$ is computed from the scene-graph annotations for the training example. We sample multiple captions per example to compute group-relative advantages for stable critic-free updates, and train on all five prompt categories for robustness across control settings.

%% file: latex/sec/figures/scope_overview.tex
\begin{figure*}[!t]
\vspace{-2mm}
  \centering
  \includegraphics[width=\textwidth]{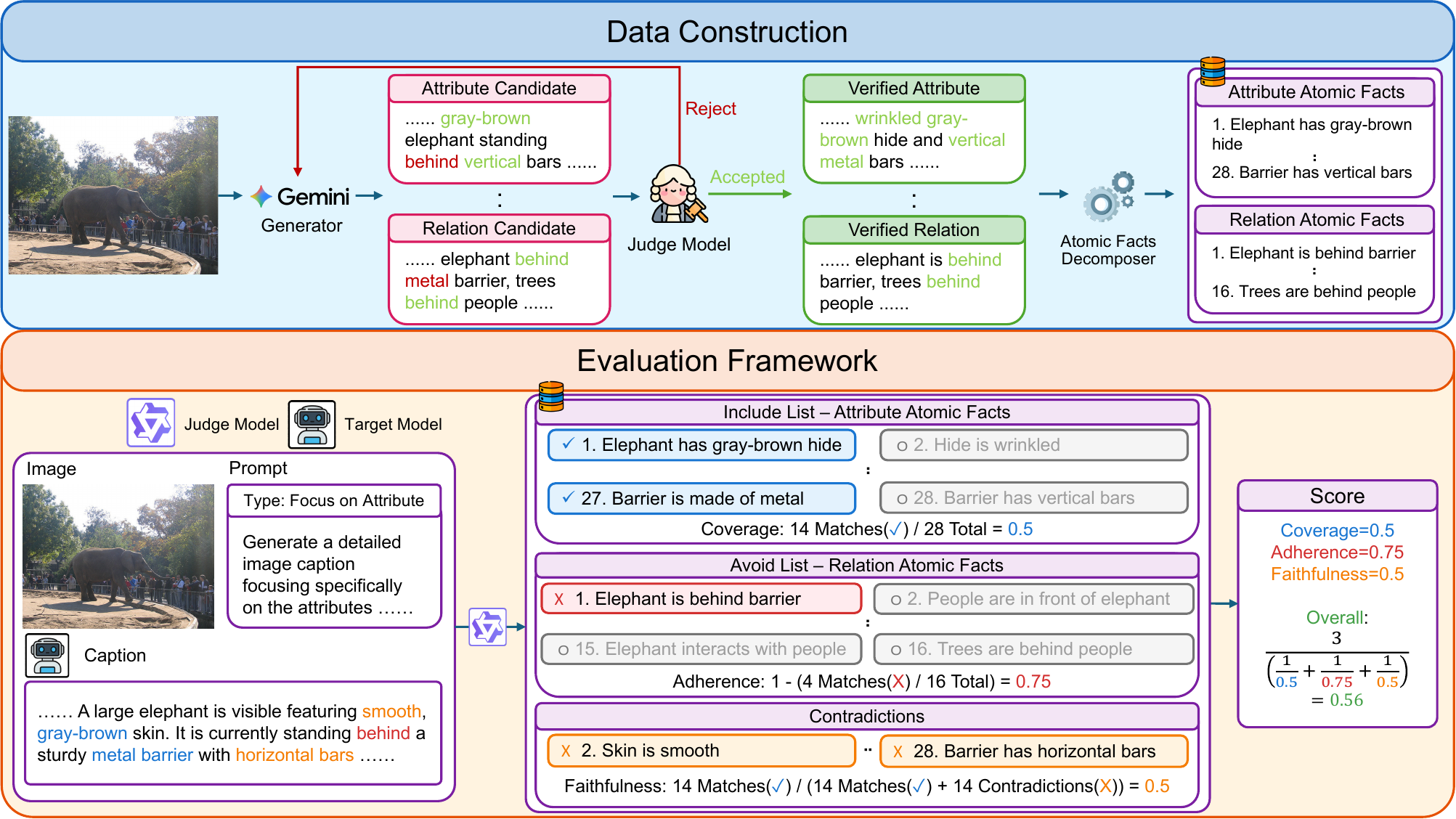}
  \caption{Overview of the \benchmark{} pipeline. The framework operates in two stages: Data Construction (top), which utilizes a generator-verifier loop to extract and decompose verified atomic facts; and the Evaluation Framework (bottom), which assesses model controllability by scoring generated captions against contrastive \textit{Include} (target) and \textit{Avoid} (out-of-scope) lists to derive Coverage, Adherence, and Faithfulness metrics.}
  \label{fig:scope}
  \vspace{-2mm}
\end{figure*}

%% file: latex/sec/5_Benchmark.tex
\section{Benchmark: \benchmark{}}
\label{sec:benchmark}

Standard captioning benchmarks measure overall quality, but not \emph{contrastive controllability}: covering target content while suppressing off-scope content. We therefore introduce \textbf{S}emantic \textbf{Co}ntrol and \textbf{P}recision \textbf{E}valuation (\textbf{\benchmark{}}), a benchmark with category-specific \emph{Include} and \emph{Avoid} lists derived from curated captions. Figure~\ref{fig:scope} illustrates the construction and evaluation pipeline.

\subsection{Benchmark Construction}

\benchmark{} is constructed in three stages: category-specific caption generation, automated refinement to enforce category focus, and atomic fact extraction to form contrastive Include/Avoid lists.

\sparagraph{Image curation.}
We manually curated 189 images from COCO, CompreCap, and DOCCI, ensuring clear category captions and no overlap with SFT or GRPO training data. See Appendix~\ref{app:scope_curation} for details.

\sparagraph{Caption generation and atomic fact extraction.}
For each image, we consider four categories: \emph{Attribute}, \emph{Relation}, \emph{Foreground}, and \emph{Background}. We use Gemini-3-Flash~\citep{google2025gemini} to generate category-specific captions, iteratively refining them to enforce a strict focus. We then decompose each verified caption into atomic facts. For an image $x$ and focus $p$, facts from the caption for $p$ form the Include list $I_{x,p}$, while facts from the complementary focus form the Avoid list $A_{x,p}$ (Attribute $\leftrightarrow$ Relation, Foreground $\leftrightarrow$ Background). Thus, both the contents and sizes of the lists vary by image and focus. This contrastive design directly measures target coverage and off-scope suppression. Details and benchmark statistics are provided in Appendix~\ref{app:scope_generation} and Appendix~\ref{app:scope_stats}.

\subsection{Evaluation Metrics}

To assess controllability on \benchmark{}, we employ an LLM judge (Qwen3-32B-Thinking~\citep{yang2025qwen3}) to semantically verify generated captions $C$ against atomic fact lists. \benchmark{} evaluates three complementary dimensions---Coverage, Adherence, and Faithfulness---and aggregates them into an overall Score. The inference prompts and LLM-judge prompts used in \benchmark{} are provided in Appendix~\ref{app:scope_inference_prompts} and Appendix~\ref{app:scope_judge_prompts}, respectively.

\sparagraph{Coverage.}
Coverage measures the recall of target information from the Include list. Let $N_{\text{match}}$ be the number of facts in $I_{x,p}$ verified in caption $C$:
\begin{equation}
\text{Coverage} = \frac{N_{\text{match}}}{|I_{x,p}|}.
\end{equation}

\sparagraph{Adherence.}
Adherence measures the caption's suppression of off-scope content. Let $N_{\text{violation}}$ be the number of facts from $A_{x,p}$ detected in $C$:
\begin{equation}
\text{Adherence} = 1 - \frac{N_{\text{violation}}}{|A_{x,p}|}.
\end{equation}

\input{latex/sec/tables/main_table}

\sparagraph{Faithfulness.}
Faithfulness measures factual consistency by penalizing cases where the caption explicitly states something incompatible with a target atomic fact. Let $N_{\text{contradiction}}$ denote the number of atomic facts in $I_{x,p}$ that are contradicted by caption $C$. We count a contradiction only when $C$ asserts an incompatible value for the same property (e.g., mismatches in color, number, material, or spatial relations), as determined by the LLM verifier; mere omissions or lack of mention are not counted as contradictions. We compute:
\begin{equation}
\text{Faithfulness} = \frac{N_{\text{match}}}{N_{\text{match}} + N_{\text{contradiction}}}.
\end{equation}

\sparagraph{Overall Score.}
We aggregate the three metrics using the harmonic mean to ensure balanced performance across all dimensions:
\begin{equation}
\text{Score} = \frac{3}{\frac{1}{\text{Coverage}} + \frac{1}{\text{Adherence}} + \frac{1}{\text{Faithfulness}}}.
\label{eq:overall_score}
\end{equation}
In Appendix~\ref{app:score_interpretation}, we discuss the interpretation of the overall score with supporting empirical evidence.

\sparagraph{Human Alignment.}
We validate \benchmark{} with an Amazon MTurk\footnote{https://www.mturk.com/} study on 140 image--category pairs. For each pair, annotators compare three captions: either a within-backbone triplet (Zero-shot, SFT, and SFT+\framework{}) or an external-model triplet (LLaVA-NeXT-34B~\citep{liu2024llavanext}, Qwen3-VL-32B-Instruct~\citep{bai2025qwen3vl}, and Gemini-3-Flash~\citep{google2025gemini}). \benchmark{} correlates strongly with human preferences (Spearman's $\rho=0.5704$), substantially outperforming a CompreCap-based controllability metric computed via category-specific reweighting ($\rho=0.2291$). These conclusions are robust to judge choice: smaller same-family and backbone-mismatched judges show similar human alignment, and fixed-judge rescoring preserves the advantage of \framework{}. Details are provided in Appendix~\ref{app:human_alignment}.

%% file: latex/sec/tables/main_table.tex
\begin{table*}[!t]
\vspace{-2mm}
\centering

\resizebox{\textwidth}{!}{

\begin{tabular}{@{}llcccccc@{}}

\toprule

\textbf{Base Model} & \textbf{Method} & \multicolumn{6}{c}{\textbf{\benchmark{}}} \\

\cmidrule(lr){3-8}

& & & \textbf{Attribute} & \textbf{Relation} & \textbf{Foreground} & \textbf{Background} & \textbf{Overall} \\

\midrule

\multirow{5}{*}{Qwen2.5-VL-3B} 

 & Zero-shot & & 5.93 & 29.66 & 13.88 & 13.17 & 15.66 \\

 & SFT & & 25.01 & 32.84 & 15.02 & 17.58 & 22.61 \\

 & SFT+CLIP & & 25.35 & 41.01 & 20.15 & 19.10 & 26.40 \\

 & SFT+CompreCap & & 30.11 & 27.84 & 20.00 & 25.46 & 25.85 \\

& \textbf{SFT+\framework{}} & & \textbf{30.44} {\footnotesize (+24.51)} & \textbf{45.55} {\footnotesize (+15.89)} & \textbf{24.38} {\footnotesize (+10.50)} & \textbf{26.58} {\footnotesize (+13.41)} & \textbf{31.74} {\footnotesize (+16.08)} \\

\midrule

\multirow{5}{*}{InternVL3-2B} 

 & Zero-shot & & 8.38 & 47.46 & 22.48 & 24.19 & 25.63 \\

 & SFT & & 32.43 & 46.28 & 20.18 & 25.93 & 31.21 \\

 & SFT+CLIP & & 27.97 & 40.38 & 23.08 & 22.11 & 28.39 \\

 & SFT+CompreCap & & 34.71 & 33.55 & 19.57 & 29.22 & 29.26 \\

& \textbf{SFT+\framework{}} & & \textbf{36.57} {\footnotesize (+28.19)} & \textbf{54.21} {\footnotesize (+6.75)} & \textbf{25.29} {\footnotesize (+2.81)} & \textbf{31.37} {\footnotesize (+7.18)} & \textbf{36.86} {\footnotesize (+11.23)} \\

\midrule

\midrule

& & \multicolumn{6}{c}{\textbf{CompreCap}} \\

\cmidrule(lr){3-8}

& & \textbf{General} & \textbf{Attribute} & \textbf{Relation} & \textbf{Foreground} & \textbf{Background} & \textbf{Overall} \\

\midrule

\multirow{5}{*}{Qwen2.5-VL-3B} 

 & Zero-shot & 41.62 & 37.17 & 47.93 & 49.90 & 38.77 & 43.08 \\

 & SFT & 41.49 & 42.59 & 50.52 & 52.53 & 48.26 & 47.08 \\

 & SFT+CLIP & 41.97 & 40.34 & 50.67 & 54.33 & 39.47 & 45.35 \\

 & SFT+CompreCap & 45.21 & 43.34 & 45.20 & 49.86 & 50.19 & 46.76 \\

& \textbf{SFT+\framework{}} & \textbf{47.51} {\footnotesize (+5.89)} & \textbf{45.05} {\footnotesize (+7.88)} & \textbf{54.68} {\footnotesize (+6.75)} & \textbf{55.66} {\footnotesize (+5.76)} & \textbf{51.20} {\footnotesize (+12.43)} & \textbf{50.82} {\footnotesize (+7.74)} \\

\midrule

\multirow{5}{*}{InternVL3-2B} 

 & Zero-shot & 39.99 & 35.33 & 50.46 & 51.24 & 38.77 & 43.16 \\

 & SFT & 44.68 & 43.79 & 52.50 & 51.95 & 48.22 & 48.23 \\

 & SFT+CLIP & 30.30 & 33.26 & 37.24 & 43.94 & 33.99 & 35.75 \\

 & SFT+CompreCap & 46.10 & 44.99 & 46.29 & 51.32 & 50.29 & 47.80 \\

& \textbf{SFT+\framework{}} & \textbf{49.38} {\footnotesize (+9.39)} & \textbf{47.25} {\footnotesize (+11.92)} & \textbf{56.38} {\footnotesize (+5.92)} & \textbf{55.38} {\footnotesize (+4.14)} & \textbf{51.67} {\footnotesize (+12.90)} & \textbf{52.02} {\footnotesize (+8.86)} \\
\bottomrule

\end{tabular}

}

\caption{Performance comparison using \benchmark{} and CompreCap. \benchmark{} and CompreCap scores for two LVLM backbones under Zero-shot, SFT, and GRPO variants. SFT+\framework{} achieves the best performance and improves category-wise control without degrading general caption quality. Parentheses denote gains over zero-shot baseline.}

\label{tab:main_table}

\vspace{-7pt}

\end{table*}

%% file: latex/sec/6_Experiments.tex
\input{latex/sec/tables/strong_prompt_overall}

\section{Experiments}
\label{sec:experiments}
\sparagraph{Experimental Setup.}
We experiment with two VLM backbones, \textbf{Qwen2.5-VL-3B-Instruct}~\citep{bai-etal-2025-qwen25vl} and \textbf{InternVL3-2B}~\citep{zhu2025internvl3}. Each model is trained with a two-stage pipeline consisting of supervised fine-tuning (SFT) followed by GRPO using the proposed prompt-conditioned control objective. Training data construction for SFT and GRPO, together with implementation details and hyperparameters, are provided in Appendix~\ref{app:exp_details_data} and Appendix~\ref{app:exp_details_impl}. For SFT+\framework{}, results are averaged over three runs with different seeds. Per-seed results and standard deviations are reported in Appendix~\ref{app:seed_stability}.

\sparagraph{Evaluation Setup.}
We use three complementary evaluation protocols:
(i) \benchmark{} to directly measure controllability via Include/Avoid lists over four categories;
(ii) a CompreCap-based component evaluation with category-specific reweighting (scaled 0--100) on the held-out CompreCap split, using stricter object validity and a stronger LLM judge (Appendix~\ref{app:exp_details_eval});
and (iii) a reference-based general-caption evaluation on 5{,}000 DOCCI test images using standard captioning metrics and fine-grained caption evaluators (Appendix~\ref{app:general_caption_eval}).

\sparagraph{Baselines.}
We compare against:
\textbf{Zero-shot} prompting of the pretrained backbone,
\textbf{SFT} (cross-entropy fine-tuning only),
\textbf{SFT+CLIP} (GRPO with CLIP similarity reward after adding CLIP-based focus indicators),
and \textbf{SFT+CompreCap} (GRPO using the original CompreCap metric as reward). The two GRPO baselines represent alternative training objectives: a holistic similarity reward and a CompreCap-style scene-graph reward. Expanded baseline descriptions are in Appendix~\ref{app:exp_details_baselines}.

%% file: latex/sec/tables/strong_prompt_overall.tex
\begin{table*}[!t]
\centering

\begin{tabular}{@{}lccccc@{}}
\toprule
\textbf{Base Model} & \multicolumn{5}{c}{\textbf{\benchmark{}}} \\
\cmidrule(lr){2-6}
& Zero-shot & SFT & SFT+CLIP & SFT+CompreCap & \textbf{SFT+\framework{}} \\
\midrule
Qwen2.5-VL-3B
& 12.80 & 23.22 & 27.37 & 27.31 & \textbf{32.56} \\
InternVL3-2B
& 20.46 & 31.01 & 26.90 & 28.50 & \textbf{35.62} \\
\bottomrule
\end{tabular}

\caption{\benchmark{} Overall scores under stronger include and exclude inference prompts.}
\label{tab:strong_prompt_overall}
\end{table*}

%% file: latex/sec/7_Result.tex
\section{Results and Analysis}
\label{sec:results_analysis}

\subsection{Main Results}
\label{subsec:main_results}
Table~\ref{tab:main_table} summarizes results on the first two evaluation protocols: \benchmark{} for contrastive controllability (Include/Avoid) and the CompreCap-based component evaluation for fine-grained factual alignment. Across two LVLM backbones, SFT+\framework{} consistently achieves the best overall performance and the strongest category-wise controllability.

\sparagraph{Controllability on \benchmark{}.}
Across both backbones, SFT+\framework{} achieves the best overall \benchmark{} score and improves consistently across all four control categories. Relative to zero-shot prompting, \framework{} yields large gains in overall controllability (approximately +16 points on Qwen2.5-VL-3B and +11 points on InternVL3-2B), indicating the learned policy follows the requested emphasis while better suppressing off-scope content. Qualitative examples in Appendix~\ref{app:focus_examples} illustrate this behavior. Notably, \framework{} also outperforms SFT and GRPO baselines (SFT+CLIP and SFT+CompreCap) on both backbones, showing that the proposed training objective yields more reliable semantic control than these alternatives. This advantage also holds when \benchmark{} is evaluated separately on CompreCap-origin and non-CompreCap images for both backbones; 95\% image-bootstrap confidence intervals on the full benchmark further show clear separation from the strongest baselines (Appendix~\ref{app:scope_robustness}). We also evaluate a stricter prompt-engineering setting with explicit inclusion/exclusion rules. As shown in Appendix~\ref{app:prompt_engineering}, stronger prompting alone does not close the gap to SFT+\framework{} on either backbone.

\sparagraph{Fine-grained factual alignment on CompreCap.}
\framework{} also delivers the strongest overall CompreCap performance for both backbones (approximately +7--9 points over zero-shot overall), with improvements that are broadly distributed across general captions and the four fine-grained categories. This indicates that the gains on \benchmark{} are accompanied by stronger compositional grounding on objects, attributes, and relations, while maintaining or improving factual correctness. These gains are consistent with our design: additional ablations in Appendix~\ref{app:weighting_ablation} show that replacing prompt-conditioned contrastive weighting with either a fixed global reward mixture or target-only up-weighting leads to weaker cross-category controllability. We additionally evaluate general captioning on 5{,}000 DOCCI test images. Across the two backbones, \framework{} largely maintains standard reference-based caption quality while improving general-prompt fine-grained alignment. Full results are reported in Appendix~\ref{app:general_caption_eval}.

\sparagraph{Control under explicit prompts.}
Prompt specificity can itself affect controllability. We therefore re-evaluate all methods using explicit inference prompts that specify both the requested semantic focus and the content to avoid, while keeping each model checkpoint fixed. As shown in Table~\ref{tab:strong_prompt_overall}, their effect is mixed: they modestly improve several methods on Qwen2.5-VL-3B, but do not improve performance on InternVL3-2B, and zero-shot performance decreases on both backbones. Even with these engineered prompts, SFT+\framework{} remains the best-performing method. This suggests that explicit prompt specification alone does not provide consistent semantic control, whereas the advantage of our learned control objective is robust to prompt formulation. Full category-wise results are provided in Appendix~\ref{app:prompt_engineering}.

\subsection{Token Efficiency Ablation}
\label{subsec:token_ablation}
Figure~\ref{fig:token_ablation} compares caption length (measured in tokens) against \benchmark{} Overall score. Zero-shot generations are the most verbose (${\sim}168$ tokens) yet achieve the lowest \benchmark{} score, consistent with producing extraneous, off-scope details that degrade controllability and precision. In contrast, SFT markedly reduces output length (${\sim}92$ tokens) but tends toward excessive brevity, limiting coverage and capping the overall score.

Our framework learns a more effective length--content trade-off: it generates moderately long captions (${\sim}110$ tokens), reducing token count by approximately 34\% relative to zero-shot while achieving the highest \benchmark{} Overall (a 43.8\% improvement). This result also clarifies why simply optimizing for shorter captions (e.g., SFT+CLIP) does not guarantee improved \benchmark{} performance: effective controllable captioning requires allocating tokens to the \emph{relevant} visual content rather than maximizing either verbosity or brevity.

\input{latex/sec/figures/token}

\subsection{Component-Scoring Ablation}
\label{subsec:ablation_reward}

Table~\ref{tab:components} ablates the three key design choices in our scene reward pipeline by toggling each component during GRPO training. Starting from the baseline configuration (no thresholding, no CoT judging, and the original Llama3-8B judge), InternVL3-2B attains an overall score of 29.26. Introducing a validity threshold for object matching yields a modest but consistent gain (+0.53), indicating that stricter matching slightly reduces reward noise from spurious semantic alignments. In contrast, enabling CoT-based verification for attribute and relation scoring produces a substantially larger improvement (+4.79), supporting our claim that stronger, reasoning-based judging is crucial to prevent overly lenient rewards for generic descriptions. Replacing the judge with Qwen3-30B-A3B also yields a similarly large gain (+5.06), demonstrating that a more capable evaluator meaningfully improves the reward quality used for policy optimization.

Combining components further increases performance: CoT combined with the Qwen judge reaches 36.67 (+7.41), and enabling all three achieves the best result of 36.86 (+7.60, representing a 26.0\% relative improvement). Overall, the ablation confirms that our framework's main gains come from strengthening attribute and relation verification via CoT prompting and using a more capable judge model, while thresholded object matching plays a smaller but complementary role in stabilizing grounding.

\input{latex/sec/tables/components}

\subsection{Hyperparameter Analysis}
\label{subsec:hyperparameter_analysis}

Figure~\ref{fig:hyper-parameters} shows the sensitivity of InternVL3-2B trained with \framework{} to two reward-computation hyperparameters: (i) Evaluation LLM used to score Attribute and Relation components, and (ii) object-matching threshold $\tau$ for the object score. We defer analysis of the negative penalty magnitude used to suppress off-scope content to Appendix~\ref{app:penalty_sensitivity}.

\sparagraph{Evaluation LLM.}
Across a diverse set of judges, \framework{} is largely stable: changing the evaluator shifts absolute \benchmark{} scores moderately, with Overall scores spanning roughly 34.6--36.9. Nevertheless, evaluator capacity matters: stronger judges provide cleaner training signals and consistently improve performance. In our experiments, Gemma3-12B~\citep{team2025gemma} and Qwen3-30B-A3B yield the highest Overall scores, while smaller judges (e.g., Qwen3-4B) underperform. The sensitivity is most pronounced for Relation, which varies the most across evaluators and largely drives the Overall trend; Foreground remains the most challenging category under all judges.

\input{latex/sec/figures/hyper-parameters}
\sparagraph{Object-matching threshold $\tau$.}
We vary the strictness of object matching via the validity threshold $\tau$. Performance remains stable across a broad range of values, indicating that \framework{} does not require finely tuned matching to function well. Nonetheless, $\tau=0.5$ achieves the best Overall result and offers the most balanced trade-off across categories. Lower thresholds slightly boost Attribute and Foreground scores but tend to hurt Relation (by admitting noisier object matches), whereas higher thresholds mildly improve Background at the expense of Foreground. These trends motivate our default configuration of Qwen3-30B-A3B with $\tau=0.5$.

%% file: latex/sec/figures/token.tex
\begin{figure}[!t]
  \vspace{-2mm}
  \centering
  \includegraphics[width=1.0\linewidth]{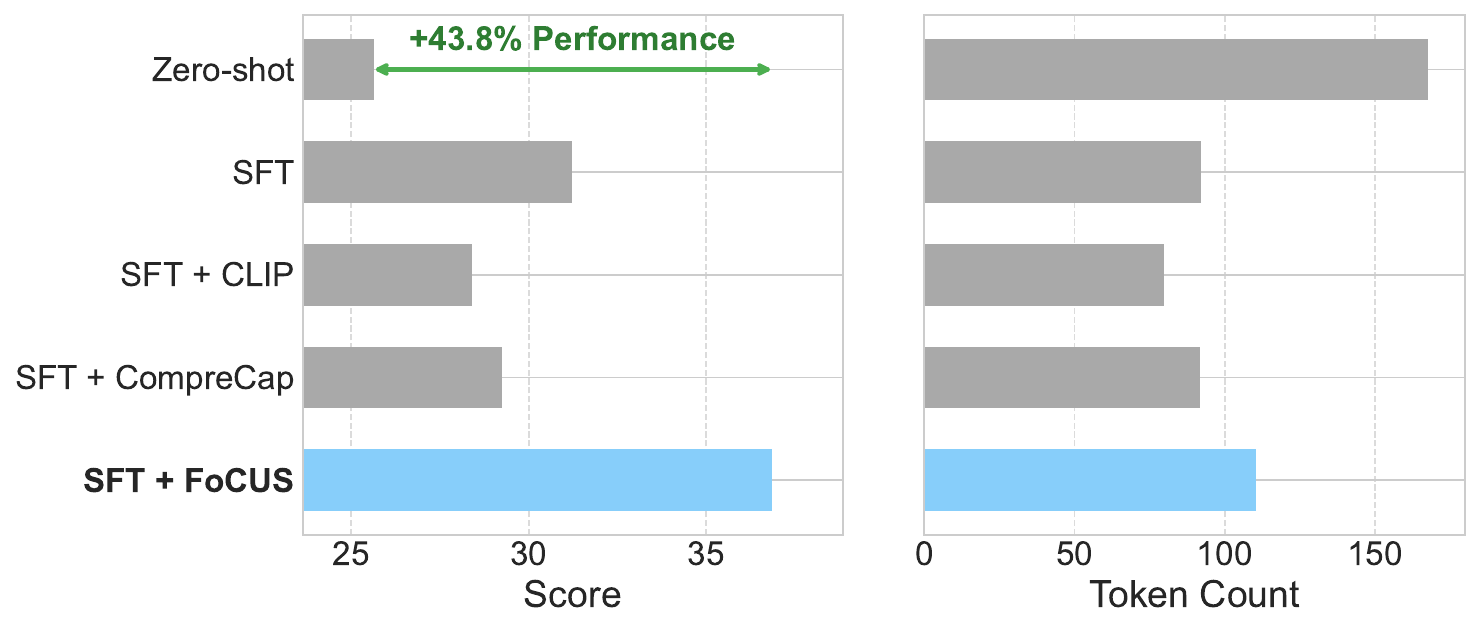}
   \caption{Token efficiency ablation: \benchmark{} Overall score (left) and average caption token count (right).}
  \label{fig:token_ablation}
  \vspace{-4mm}
\end{figure}

%% file: latex/sec/tables/components.tex
\begin{table}[!t]
\vspace{-2mm}
\centering
\resizebox{\columnwidth}{!}{
\begin{tabular}{@{}ccc|cc@{}}
\toprule
\multicolumn{3}{c|}{\textbf{Configuration}} & \textbf{Overall} & \textbf{$\Delta$ from} \\
\cmidrule(r){1-3}
\textbf{Threshold} & \textbf{CoT} & \textbf{Qwen} & \textbf{Score} & \textbf{Baseline} \\
\midrule
 &  &  & 29.26 & --- \\
\checkmark &  &  & 29.79 & $+0.53$ ($+1.8\%$) \\
 & \checkmark &  & 34.05 & $+4.79$ ($+16.4\%$) \\
 &  & \checkmark & 34.32 & $+5.06$ ($+17.3\%$) \\
\checkmark & \checkmark &  & 35.15 & $+5.89$ ($+20.1\%$) \\
\checkmark &  & \checkmark & 34.61 & $+5.35$ ($+18.3\%$) \\
 & \checkmark & \checkmark & 36.67 & $+7.41$ ($+25.3\%$) \\
\checkmark & \checkmark & \checkmark & \textbf{36.86} & $\mathbf{+7.60}$ ($\mathbf{+26.0\%}$) \\
\bottomrule
\end{tabular}
}
\caption{Ablation of validity thresholding, CoT verification, and the Qwen judge during GRPO.}
\label{tab:components}
\vspace{-4mm}
\end{table}

%% file: latex/sec/figures/hyper-parameters.tex
\begin{figure}[!t]
\vspace{-2mm}
  \centering
  \includegraphics[width=1.0\linewidth]{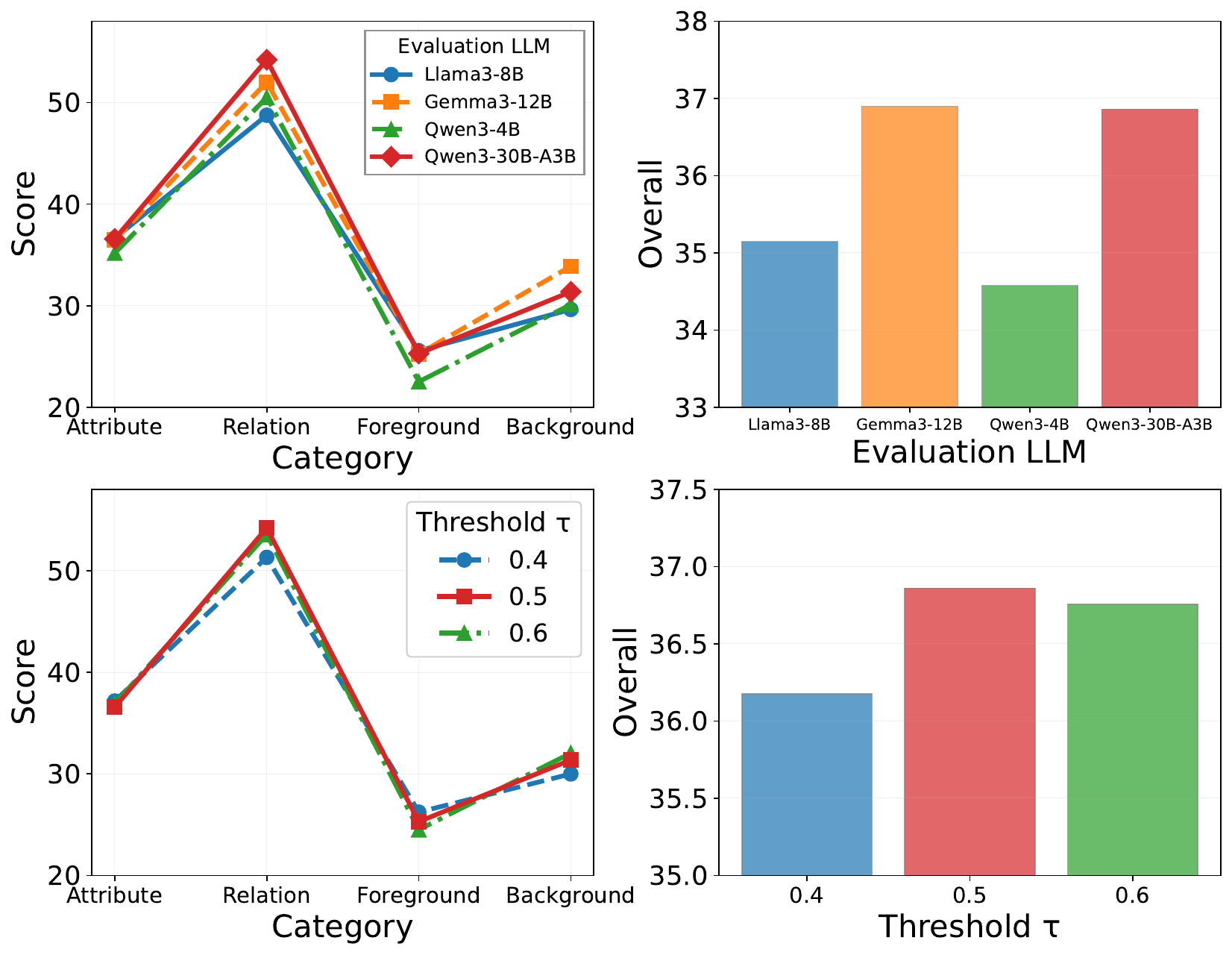}
  \caption{Hyperparameter sensitivity of \framework{}: \benchmark{} category-wise scores (left) and Overall (right) when varying the evaluation LLM (top) and object-matching threshold \(\tau\) (bottom).}
  \label{fig:hyper-parameters}
  \vspace{-10pt}
\end{figure}

%% file: latex/sec/8_Conclusion.tex
\section{Conclusion}
\label{sec:conclusion}

We presented \framework{}, a controllable image captioning method
using a prompt-conditioned, scene-graph-aligned objective to emphasize requested content and suppress off-scope content by differentially weighting scene-graph component scores according to natural-language control prompts. We also introduced \benchmark{}, a benchmark with contrastive Include/Avoid constraints for evaluating semantic control. Experiments on two VLM backbones show that \framework{} improves controllability and fine-grained caption quality without requiring architectural modifications. Together, these results suggest that prompt-conditioned scene-graph control is a practical and effective direction for controllable image captioning.

%% file: latex/sec/98_Limitations.tex
\section*{Limitations}

Our work has several limitations that suggest directions for future research. First, \framework{} relies on scene graph parsing and LLM-based evaluation during training, which introduces computational overhead and may propagate parsing errors into the reward signal. Second, while we demonstrate effectiveness on two VLM backbones, the scalability of our approach to significantly larger models remains to be validated. Third, the \benchmark{} benchmark employs LLM-based evaluation to assess Coverage, Adherence, and Faithfulness, which inherits potential biases and inconsistencies from the underlying language model judge; although we validate human alignment through an MTurk study showing reasonable correlation, LLM evaluators may still exhibit systematic blind spots or fail to capture subtle semantic distinctions that human annotators would identify.

%% file: latex/sec/99_Ethics_Statement.tex
\section*{Ethics Statement}
This research adhered to the ACL Ethics Policy, utilizing only publicly available datasets and vision-language models for training and evaluation. The \benchmark{} benchmark was constructed using publicly available images and automated generation pipelines, with human validation conducted through MTurk following standard ethical guidelines for crowdsourced annotation. The goal of this work is to improve controllability in image captioning systems, and no negative ethical outcomes are anticipated.

%% file: latex/sec/A_CompreCap.tex
\section{Background on CompreCap and Reward Design Considerations}
\label{app:comprecap}

\subsection{Overview of CompreCap}

CompreCap~\citep{lu2025benchmarking} is a fine-grained evaluation framework for image captioning based on structured scene-graph analysis. Compared with holistic metrics such as CIDEr~\citep{vedantam2015cidera} and earlier scene-graph-based metrics such as SPICE~\citep{anderson2016spice}, CompreCap decomposes both the ground-truth annotations and generated captions into three semantic components: objects, attributes, and relations. The evaluation pipeline operates in two stages:
\begin{enumerate}
    \item \textbf{Parsing:} It employs \texttt{spaCy}~\citep{honnibal2020spacy} to extract nouns (objects) and parses the caption to associate modifiers (attributes) and spatial prepositions (relations) with specific objects.
    \item \textbf{Scoring:} It computes a weighted average of component scores. Object matching is performed using Sentence-BERT embeddings, while attributes and relations are evaluated using a Large Language Model (LLM) to judge semantic alignment.
\end{enumerate}
Although CompreCap is effective for post-hoc evaluation, our preliminary experiments indicate that directly using its score as a reward signal introduces significant noise due to two structural limitations.

\subsection{Practical Considerations When Using CompreCap as a Learning Signal}

\paragraph{Semantic Drift in Object Scoring.}
CompreCap computes object scores from embedding similarity, which can assign moderate similarity to semantically related but incorrect pairs (e.g., matching \texttt{house} to \texttt{environment}). When such pairs are treated as matches, downstream attribute and relation scoring may be attached to the wrong entity, which adds noise to the learning signal. To reduce this effect, we use mutual-best matching together with a validity threshold ($\tau=0.5$), as described in Section~\ref{sec:method}.

\paragraph{Leniency in Attribute and Relation Scoring.}
Standard zero-shot LLM judges (e.g., Llama3-8B~\citep{grattafiori2024llama}) can be lenient for attribute and relation scoring, often assigning high scores to generic descriptions that omit requested specificity. For example, a caption containing only ``refrigerator'' may receive substantial credit against a ground-truth description such as ``white fridge with side-by-side doors.'' As a training signal, this weakens pressure to produce precise descriptions. We therefore use Chain-of-Thought prompting with a stronger reasoning model (\texttt{Qwen3-30B-A3B-Instruct}) to obtain more discriminative attribute and relation scores.

%% file: latex/sec/A_Object_Matching.tex
\section{Object Matching Details}
\label{app:object_matching}

\subsection{Scene Graph Parsing}

Given a generated caption $y$ and ground-truth scene graph annotations, we parse the caption into constituent scene graph elements. We employ the spaCy~\citep{honnibal2020spacy} parser for noun extraction and lemmatization, decomposing the generated caption into sentences and extracting candidate objects. For each identified object, we associate the relevant sub-captions that mention it, enabling object-bound attribute and relation evaluation.

\subsection{Mutual-Best Matching with Validity Threshold}

Let $\mathcal{C}=\{c_i\}_{i=1}^{n}$ be the candidate object nouns extracted from the generated caption, and let $\mathcal{G}=\{g_j\}_{j=1}^{m}$ be the ground-truth object categories from the annotated scene graph. We use Sentence-BERT~\citep{reimers2019sentencebert} with the
all-MiniLM-L6-v2 pretrained model and compute a similarity matrix:
\begin{equation}
\mathbf{S}\in\mathbb{R}^{n\times m}, \quad S_{i,j}=\cos\!\bigl(\phi(c_i),\phi(g_j)\bigr)
\end{equation}
where $\phi(\cdot)$ denotes the Sentence-BERT embedding.

To obtain discrete matches while preventing semantic drift, we construct a mutual-best match indicator matrix $\mathbf{S}'\in\{0,1\}^{n\times m}$:
\begin{equation}
\begin{split}
S'_{i,j} = \mathbb{I}\bigl[
  & j = \argmax_{j'} S_{i,j'} \;\wedge\; i = \argmax_{i'} S_{i',j} \\
  & \wedge\; S_{i,j} \ge \tau \bigr],
\end{split}
\label{eq:mutual_best_match}
\end{equation}
where $\mathbb{I}[\cdot]$ is the indicator function and $\tau$ is a validity threshold. In our experiments, we set $\tau=0.5$. The mutual-best constraint ensures each ground-truth object $g_j$ matches at most one candidate $c_i$, so $\sum_i S'_{i,j}\in\{0,1\}$.

This formulation prevents partial credit for semantically distant matches caused by embedding-space artifacts, yielding a cleaner object-level signal for GRPO optimization.

\section{Prompt Category Descriptions}
\label{app:prompt_categories}

We use five control prompt categories for controllable caption generation:

\begin{itemize}
\item \textbf{General Detailed Caption} instructs the model to generate a comprehensive description covering all visual content.
\item \textbf{Attribute-Focused Caption} emphasizes detailed attribute descriptions of objects (colors, textures, shapes, materials, states, etc.).
\item \textbf{Relation-Focused Caption} prioritizes spatial and functional relationships between objects.
\item \textbf{Foreground-Focused Caption} concentrates on main subjects and salient objects.
\item \textbf{Background-Focused Caption} emphasizes contextual elements, settings, and peripheral content.
\end{itemize}

\section{Complete Reward Weight Configurations}
\label{app:reward_weights}

For each prompt category, we use the following prompt-conditioned weights to align the control objective with the requested semantic emphasis:

\paragraph{General Caption.}
\begin{equation}
R_\text{general} = 0.25 \tilde{S}_\text{obj} + 0.35 \tilde{S}_\text{attr} + 0.40 \tilde{S}_\text{rel}
\end{equation}
where $\tilde{S}$ denotes scores linearly scaled to $[0, 1]$.

\paragraph{Attribute-Focused Caption.}
\begin{equation}
R_\text{attr} = 0.1 \cdot \tilde{S}_\text{obj} + 0.9 \cdot \tilde{S}_\text{attr} - 1.0 \cdot \tilde{S}_\text{rel}
\end{equation}
The negative weight on relations discourages off-scope relational content while keeping the objective centered on object properties.

\paragraph{Relation-Focused Caption.}
\begin{equation}
R_\text{rel} = 0.1 \cdot \tilde{S}_\text{obj} - 1.0 \cdot \tilde{S}_\text{attr} + 0.9 \cdot \tilde{S}_\text{rel}
\end{equation}
The negative weight on attributes analogously discourages off-scope attribute content when the requested emphasis is relational.

\paragraph{Foreground-Focused Caption.}
\begin{equation}
R_\text{fg} = \tilde{S}_\text{fg} - \tilde{S}_\text{bg}
\end{equation}
This formulation rewards captions that emphasize main subjects while de-emphasizing background content.

\paragraph{Background-Focused Caption.}
\begin{equation}
R_\text{bg} = \tilde{S}_\text{bg} - \tilde{S}_\text{fg}
\end{equation}

%% file: latex/sec/A_SCOPE_Construction.tex
\section{\benchmark{} Construction Details}
\label{app:scope_construction}
\input{latex/sec/tables/A_pass-rate}

\subsection{Image Curation}
\label{app:scope_curation}

For category-specific caption generation, we manually curated 189 images drawn from three established datasets: COCO (142 images), CompreCap (35 images), and DOCCI (12 images). All three datasets are licensed under Creative Commons Attribution 4.0. Care was taken to ensure that the selected images do not overlap with those used in the SFT and GRPO training sets.

\subsection{Caption Generation and Fact Extraction}
\label{app:scope_generation}

\paragraph{Category-specific candidate generation.}
We employ Gemini-3-Flash~\citep{google2025gemini} as the caption generator. For each image, we define four semantic categories: Attribute, Relation, Foreground, and Background. For each category, we prompt the model to generate five diverse candidate captions with instructions to focus exclusively on the target aspect (e.g., ``Describe only the foreground objects without mentioning the background context'').

\input{latex/sec/tables/A_scope_facts_lists}

\paragraph{Automated quality assurance via self-refinement.}

To reduce out-of-scope content in category captions (e.g., background details appearing in a foreground-focused caption), we use Gemini-3-Flash as a judge within a generator--verifier loop with hard category-focus constraints. Each candidate is checked for strict adherence to the requested category. Captions containing out-of-scope information are rejected and regenerated for up to three attempts; if no candidate passes after exhausting this budget, the generation process for that image--category pair is restarted from scratch. We retain only verifier-approved captions in the final benchmark.

These pass rates therefore quantify the success of the iterative construction pipeline, rather than the model's single-shot controllability. Across the 756 image--category construction instances ($189 \times 4$), 92.1\% (696/756) obtained a verifier-approved caption within three attempts before a full restart was needed. The first-attempt pass rate is substantially lower and varies by category: 17.5\% for Attribute, 69.3\% for Relation, 19.6\% for Foreground, and 46.0\% for Background, or 38.1\% overall (Table~\ref{tab:scope_pass_rate}). This gap highlights that strict category adherence is achieved primarily through iterative verification and regeneration.

\paragraph{Atomic fact extraction and contrastive lists.}
We decompose verified captions into atomic facts using the same LLM. For a given category, its atomic facts form the Include list used during evaluation. We do not manually construct negative examples; instead, the Avoid list is derived from paired categories. For instance, the Include list for Attribute serves as the Avoid list when evaluating Relation, enabling direct measurement of the model's ability to emphasize target content while suppressing complementary information.

%% file: latex/sec/tables/A_pass-rate.tex
\begin{table}[t]
\centering
\resizebox{\columnwidth}{!}{
\begin{tabular}{lccccc}
\toprule
Metric & Attribute & Relation & Foreground & Background & Overall \\
\midrule
Pass rate & 17.5\% & 69.3\% & 19.6\% & 46.0\% & 38.1\% \\
\bottomrule
\end{tabular}
}
\caption{First-attempt pass rates for the Gemini-3-Flash generator--verifier loop used in \benchmark{} construction. Rates are computed before any regeneration.}
\label{tab:scope_pass_rate}
\end{table} 

%% file: latex/sec/tables/A_scope_facts_lists.tex
\begin{table*}[!t]
\centering
\resizebox{\textwidth}{!}{
\begin{tabular}{lccccc}
\toprule
 Statistic & Attribute & Relation & Foreground & Background & Avg. across categories \\
\midrule
Mean \#Facts $\pm$ Std & 28.77 $\pm$ 6.23 & 16.53 $\pm$ 3.53 & 18.35 $\pm$ 4.33 & 13.73 $\pm$ 3.06 & 19.35 $\pm$ 2.90 \\
\bottomrule
\end{tabular}
}
\caption{Average number of atomic facts per image in \benchmark{} by category. The final column reports the average over the four category-specific fact lists.}
\label{tab:scope_fact_stats}
\end{table*}

%% file: latex/sec/A_SCoPE-Stats.tex
\section{\benchmark{} Statistics}
\label{app:scope_stats}

\subsection{Number of Atomic Facts by Category}
\label{app:scope_fact_stats}

\benchmark{} does not impose a fixed upper bound on the number of atomic facts per category or per image, so fact-list lengths vary across categories and images. Table~\ref{tab:scope_fact_stats} reports the mean number of atomic facts per category per image, together with standard deviations. Attribute-focused lists are the longest on average (28.77 $\pm$ 6.23 facts), followed by foreground (18.35 $\pm$ 4.33), relation (16.53 $\pm$ 3.53), and background (13.73 $\pm$ 3.06). Averaged across categories, \benchmark{} contains 19.35 $\pm$ 2.90 atomic facts per category per image. Equivalently, aggregating the four category-specific lists yields approximately 77.38 atomic facts per image on average. These statistics confirm that \benchmark{} typically contains dozens of atomic facts per image.

\subsection{Object Frequency}
\label{app:scope_object_frequency}

We analyze the object vocabulary extracted from the curated atomic facts. Across the full benchmark, \benchmark{} contains 1,290 unique object types, indicating broad semantic coverage beyond a small set of recurring entities. The top-5 most frequent object types are trees (74), sky (65), building (46), grass (43), and sign (34). These statistics reflect broad object coverage and substantial diversity.

\input{latex/sec/tables/A_scope_scene_coverage}
\subsection{Scene Coverage}
\label{app:scope_scene_coverage}

We measure how much of each image is covered by the objects mentioned in \benchmark{}. Following the benchmark construction in Section~\ref{sec:benchmark}, we aggregate object mentions extracted from the curated atomic facts, ground them with OWLv2 (owlv2-large-patch14-ensemble)~\cite{NEURIPS2023_e6d58fc6}, and segment them with SAM2.1 (sam2.1-hiera-large)~\cite{ravi2025sam}. We define Scene Coverage as the fraction of image area covered by the union of the resulting masks.

Table~\ref{tab:scope_scene_coverage} shows that the object mentions in \benchmark{} cover most of the scene across all three source datasets. Applying the same grounding-and-segmentation pipeline to the original reference captions yields lower coverage overall, especially for COCO and DOCCI. This suggests that the category-specific captions and atomic facts used in \benchmark{} capture a broad portion of the visible scene, rather than only a small set of salient objects.

%% file: latex/sec/tables/A_scope_scene_coverage.tex
\begin{table}[!t]
\centering
\resizebox{\columnwidth}{!}{
\begin{tabular}{lccc}
\toprule
Object source & COCO & CompreCap & DOCCI \\
\midrule
Original reference captions & 48.89 & 83.36 & 74.09 \\
\benchmark{} atomic facts & \textbf{83.21} & \textbf{90.32} & \textbf{84.66} \\
\bottomrule
\end{tabular}
}
\caption{Average Scene Coverage (\%). Object mentions are grounded with OWLv2 and segmented with SAM2.1, with coverage computed from the union of object masks.}
\label{tab:scope_scene_coverage}
\end{table}

%% file: latex/sec/A_overall_interpre.tex
\input{latex/sec/tables/A_overall_interpre_caption}

\section{Interpreting the Overall \benchmark{} Score}
\label{app:score_interpretation}

The overall \benchmark{} score is designed to reward balanced controllability. Coverage measures inclusion of requested content, Adherence measures suppression of complementary off-scope content, and Faithfulness measures factual consistency. By combining these three dimensions with the harmonic mean (Eq.~\ref{eq:overall_score}), \benchmark{} favors captions that add relevant, grounded details while maintaining semantic focus.

This behavior is reflected in the token-length analysis in Figure~\ref{fig:token_ablation}. The shortest method, SFT+CLIP (79.74 tokens), does not achieve the best overall score, while the most verbose model, Zero-shot (167.57 tokens), performs worst. The strongest model, SFT+\framework{}, achieves the best overall result at an intermediate length (110.38 tokens), indicating that performance is driven by how effectively tokens are allocated to on-scope content rather than by brevity or verbosity alone.

Table~\ref{tab:score_interpretation_case} shows the same pattern on the foreground-focused example from Figure~\ref{fig:focus_example1}. Compared with SFT+CompreCap, the SFT+\framework{} caption is longer, but its additional details remain foreground-relevant and visually grounded, leading to a substantially higher \benchmark{} score. Together, these results suggest that higher \benchmark{} scores reflect more selective and semantically focused captioning.

%% file: latex/sec/tables/A_overall_interpre_caption.tex
\begin{table*}[t]
\centering
\small
\resizebox{\textwidth}{!}{

\begin{tabular}{p{2.8cm}p{10.3cm}cc}
\toprule
Method & Caption & Tokens & \benchmark{} Overall \\
\midrule
SFT + CompreCap & \emph{A woman in a black sleeveless dress and large black hoop earrings stands holding a smartphone, her face illuminated by the phone's screen.} & 27 & 15.00 \\
\midrule
SFT + \framework{} & \emph{A young woman in a black sleeveless dress and large black hoop earrings stands intently looking at a smartphone she holds in both hands, wearing a headband and a glowing neon-purple waistband.} & 39 & 28.57 \\
\bottomrule
\end{tabular}
}
\caption{Example-level comparison on the foreground-focused image in Figure~\ref{fig:focus_example1}. The higher-scoring \framework{} caption uses additional tokens for on-scope, visually grounded details.}
\label{tab:score_interpretation_case}
\end{table*}

%% file: latex/sec/A_Human_Alignment.tex
\section{Human Alignment Study and Judge Robustness}
\label{app:human_alignment}
\input{latex/sec/figures/MTurk_UI}

\input{latex/sec/tables/A_human_alignment}

\subsection{Human Evaluation Protocol}
We evaluate how well automatic controllability metrics align with human judgments on a subset of \benchmark{}. We construct 140 image–category pairs from the 35 images overlapping between \benchmark{} and CompreCap, spanning four control categories: \textit{Attribute}, \textit{Relation}, \textit{Foreground}, and \textit{Background}. For each image--category pair, annotators compare exactly three captions. The three captions are either the Zero-shot/SFT/SFT+\framework{} outputs from a single backbone (Qwen2.5-VL-3B-Instruct or InternVL3-2B) or the outputs of three external LVLMs (LLaVA-NeXT-34B~\citep{liu2024llavanext}, Qwen3-VL-32B-Instruct~\citep{bai2025qwen3vl}, and Gemini-3-Flash~\citep{google2025gemini}). We then collect human judgments on Amazon Mechanical Turk (MTurk). 
We restrict participation to MTurk workers with at least 10{,}000 approved HITs and a $\ge$98\% approval rate, and we provide in-task instructions defining the evaluation criteria (Recall/Precision/Accuracy) and the required 0--5 scoring plus forced ranking procedure (Figures~\ref{fig:mturk_instructions}--\ref{fig:mturk_interface}). For each image--category pair, 14 independent annotators provide scores and rankings for the three captions.

\subsection{Metrics Compared and Correlation Analysis}
For each image--category pair, we compute a metric-induced ordering of the three candidate captions and compare it against the aggregated human preference ordering. We report rank-correlation using Kendall's $\tau$ and Spearman's $\rho$ (with two-sided $p$-values). In addition to \benchmark{}, we also evaluate a CompreCap-based controllability score, where we assign different component weights per control category to measure controllability (i.e., different weightings for Attribute/Relation/Foreground/Background emphasis).

\input{latex/sec/tables/A_judge_alignment_additional}
\input{latex/sec/tables/A_scope_fixed_judges}
\subsection{Human Alignment Results}

Table~\ref{tab:human_alignment} reports the correlations between automatic metrics and human judgments. \benchmark{} shows substantially stronger and statistically significant alignment with human preferences than the CompreCap controllability-weighting variant. In particular, \benchmark{} achieves Kendall's $\tau=0.5245$ and Spearman's $\rho=0.5704$, compared with $0.2048$ and $0.2291$ for CompreCap. All correlations are statistically significant ($p<10^{-4}$). These results show that \benchmark{} aligns substantially better with human judgments than CompreCap in a candidate pool that includes the (Zero-shot, SFT, SFT+\framework{}) triplets for Qwen2.5-VL-3B-Instruct and InternVL3-2B, alongside LLaVA-NeXT-34B, Qwen3-VL-32B-Instruct, and Gemini-3-Flash. This supports our claim that the contrastive Include/Avoid formulation better reflects the controllability judgments made by human evaluators than CompreCap-style component reweighting.

\subsection{Robustness to Judge Choice}
\label{app:judge_robustness}

Because both \benchmark{} and our training reward rely on LLM judges, we additionally test whether the conclusions in the main text depend on a particular evaluator. We study two questions: (i) whether the human alignment reported in Table~\ref{tab:human_alignment} is stable across judge families and model sizes, and (ii) whether the gains of SFT+\framework{} persist when \benchmark{} is rescored with fixed weaker or backbone-mismatched judges.

\paragraph{Judge--human alignment across families and sizes.}
Using the same MTurk preference set described above, we recompute rank correlation with three additional judges beyond the default Qwen3-32B-Thinking: a much smaller same-family model (Qwen3-4B-Thinking-2507), the exact training-time judge used in our reward pipeline (Qwen3-30B-A3B-Instruct-2507), and an unseen backbone (GLM-Flash~\cite{team2025glm45}).
Table~\ref{tab:judge_alignment_additional} shows that all four judges remain strongly and statistically significantly correlated with human preferences, with only modest variation across evaluator family and size. Pairwise agreement among judges is also consistently high (Spearman's $\rho$ between 0.64 and 0.73), indicating that \benchmark{} is not narrowly tied to one evaluator's preferences.

\paragraph{Choice of the default \benchmark{} judge.}
These results motivate our use of Qwen3-32B-Thinking as the default \benchmark{} evaluator. Among all judges tested, it achieves the highest agreement with MTurk preferences, making it the strongest empirical proxy for human judgments on \benchmark{}'s Include/Avoid and contradiction-verification tasks. This evaluator is also distinct from the training-time reward judge, Qwen3-30B-A3B-Instruct-2507, reducing concern that evaluation simply reuses the optimization signal. The fixed-judge rescoring below further verifies that the main conclusions do not depend on this default choice.

\paragraph{Fixed-judge rescoring.}
We next rescore the InternVL3-2B results from Table~\ref{tab:main_table} using three fixed evaluators: the training-time reward judge (Qwen3-30B-A3B-Instruct-2507), a substantially smaller same-family judge (Qwen3-4B-Thinking-2507), and a backbone-mismatched judge (GLM-Flash). As shown in Table~\ref{tab:scope_fixed_judges}, SFT+\framework{} achieves the best \benchmark{} Overall score under all three evaluators. Thus, the gains reported in the main text are not explained by matching a single judge's style; they persist under weaker and architecturally different judges, supporting the conclusion that \framework{} improves semantic controllability itself.

%% file: latex/sec/figures/MTurk_UI.tex
\begin{figure*}[p]  \centering
  \includegraphics[width=\textwidth]{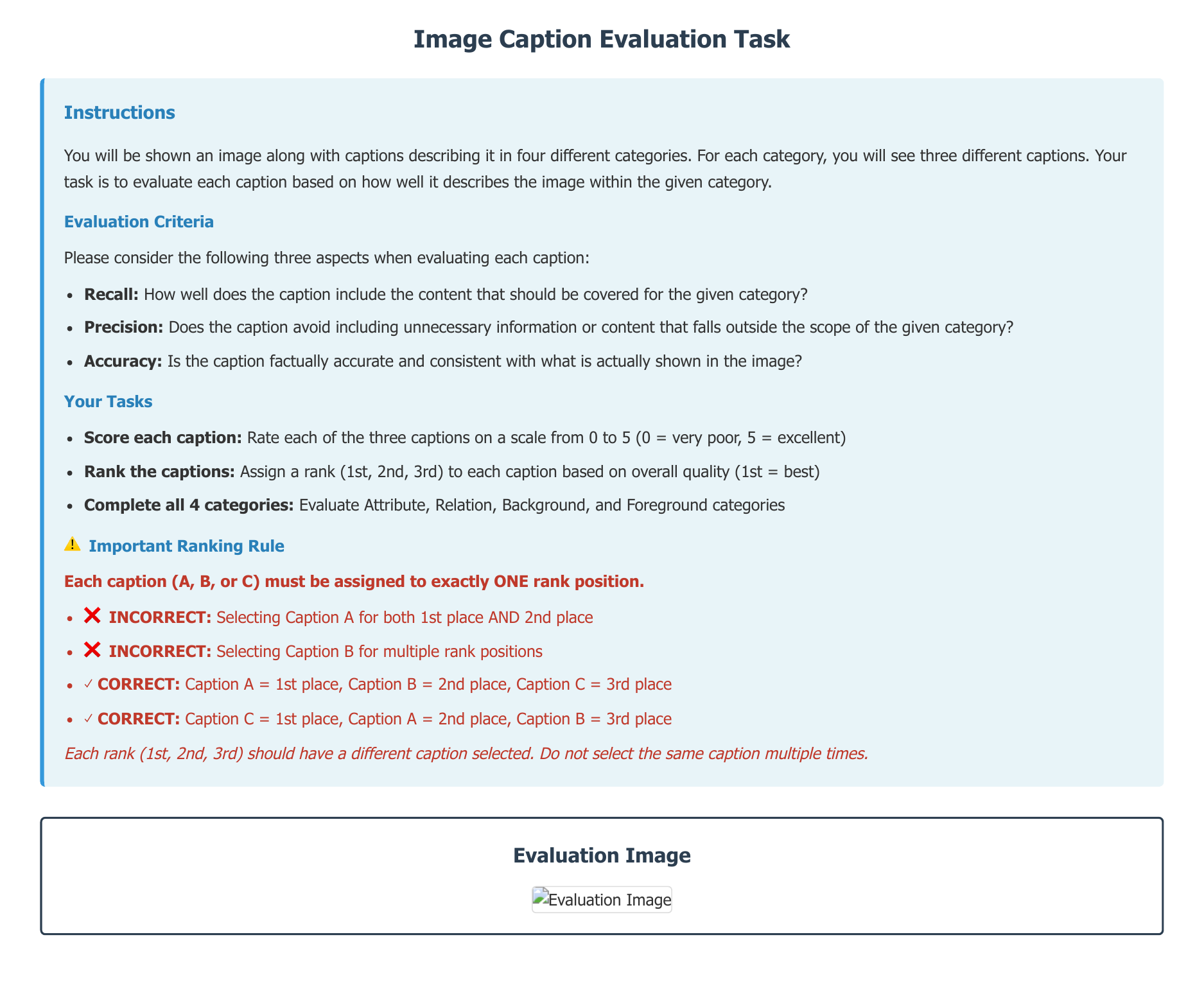}
  \caption{MTurk task instructions shown to annotators for the Image Caption Evaluation Task, including the definitions of Recall, Precision, and Accuracy, and the required workflow (rate each caption from 0 to 5 and then rank the three captions with a unique 1st/2nd/3rd choice).}
  \label{fig:mturk_instructions}
\end{figure*}

\begin{figure*}[p]  \centering
  \includegraphics[width=0.9\textwidth]{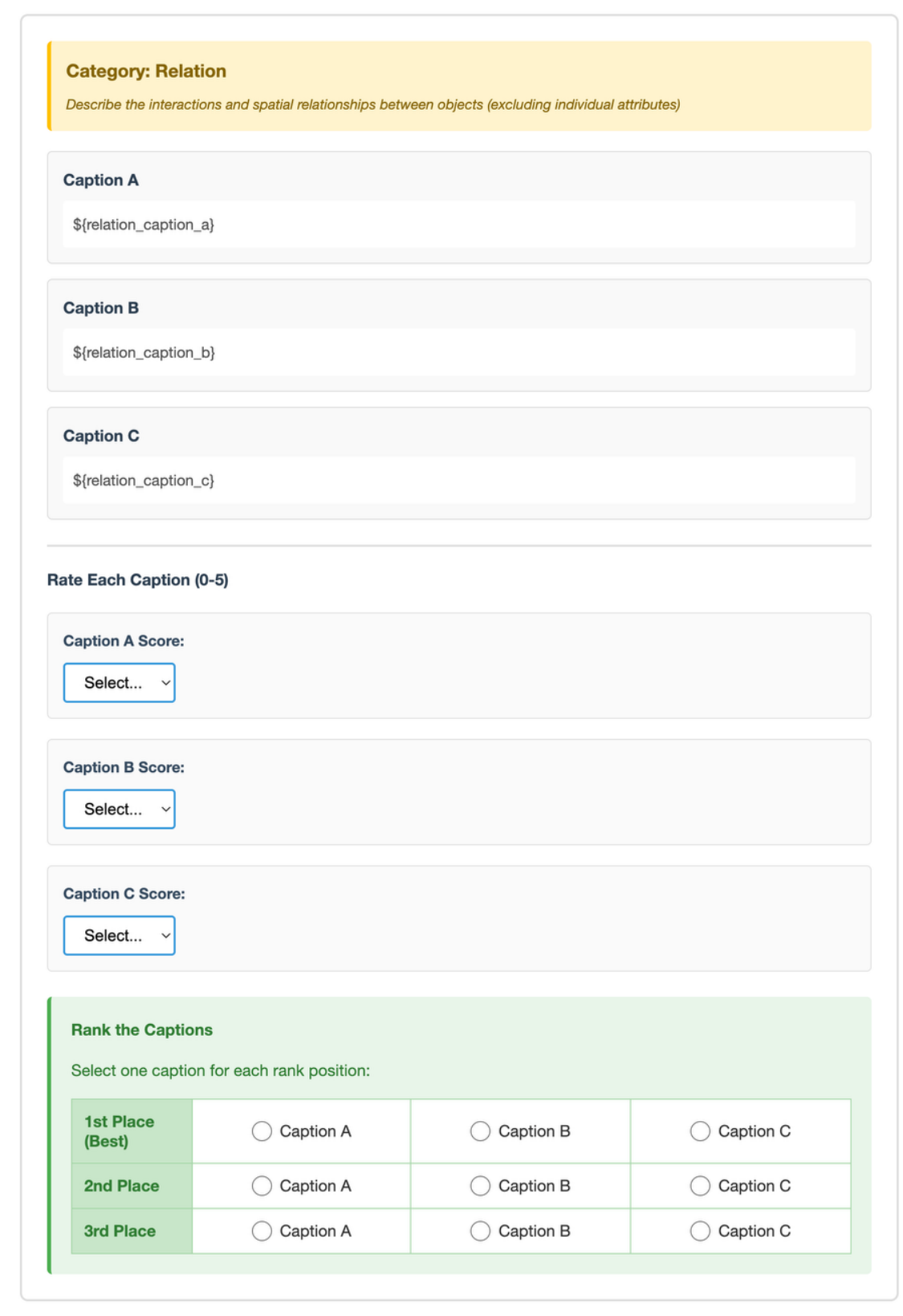}
  \caption{MTurk annotation interface for one category (Relation shown): annotators view three candidate captions (A/B/C), assign each a 0–5 score, and provide a forced ranking (1st/2nd/3rd) with no ties. The category header and guidance text change depending on whether the question targets Attribute, Relation, Foreground, or Background.}
  \label{fig:mturk_interface}
\end{figure*}

%% file: latex/sec/tables/A_human_alignment.tex
\begin{table}[!t]
\centering
\resizebox{\columnwidth}{!}{
\begin{tabular}{lcccc}
\toprule
\textbf{Metric} & \textbf{Kendall's $\tau$} & \textbf{$p$-value ($\tau$)} & \textbf{Spearman's $\rho$} & \textbf{$p$-value ($\rho$)} \\
\midrule
CompreCap & 0.2048 & $1.23{\times}10^{-5}$ & 0.2291 & $1.10{\times}10^{-5}$ \\
\benchmark{} & \textbf{0.5245} & $\mathbf{1.02{\times}10^{-7}}$ & \textbf{0.5704} & $\mathbf{8.54{\times}10^{-8}}$ \\
\bottomrule
\end{tabular}
}
\caption{Human alignment on \benchmark{}.}
\label{tab:human_alignment}
\end{table}

%% file: latex/sec/tables/A_judge_alignment_additional.tex
\begin{table*}[!t]
\centering
\begin{tabular}{lcccc}
\toprule
\textbf{Model Name} & \textbf{Kendall's $\tau$} & \textbf{$p$-value ($\tau$)} & \textbf{Spearman's $\rho$} & \textbf{$p$-value ($\rho$)} \\
\midrule
Qwen3-32B-Thinking & \textbf{0.5245} & $\mathbf{1.02{\times}10^{-7}}$ & \textbf{0.5704} & $\mathbf{8.54{\times}10^{-8}}$ \\
Qwen3-4B-Thinking & 0.5123 & $1.48{\times}10^{-5}$ & 0.5634 & $1.43{\times}10^{-5}$ \\
Qwen3-30B-A3B-Instruct & 0.4755 & $1.76{\times}10^{-7}$ & 0.5218 & $1.19{\times}10^{-7}$ \\
GLM-Flash & 0.4979 & $8.70{\times}10^{-7}$ & 0.5523 & $5.60{\times}10^{-7}$ \\
\bottomrule
\end{tabular}
\caption{Judge--human alignment on the MTurk preference set.}
\label{tab:judge_alignment_additional}
\end{table*}

%% file: latex/sec/tables/A_scope_fixed_judges.tex
\begin{table*}[!t]
\centering

\resizebox{\textwidth}{!}{
\begin{tabular}{@{}llccccc@{}}
\toprule
\textbf{Evaluation Judge} & \textbf{Method} & \multicolumn{5}{c}{\textbf{\benchmark{}}} \\
\cmidrule(lr){3-7}
& & \textbf{Attribute} & \textbf{Relation} & \textbf{Foreground} & \textbf{Background} & \textbf{Overall} \\
\midrule

\multirow{5}{*}{Qwen3-30B-A3B-Instruct}
 & Zero-shot & 16.68 & 61.92 & 33.23 & 39.74 & 37.89 \\
 & SFT & 47.45 & 59.52 & 30.77 & 49.64 & 46.85 \\
 & SFT+CLIP & 41.36 & 54.74 & 33.71 & 39.92 & 42.43 \\
 & SFT+CompreCap & 48.63 & 44.53 & 29.00 & 52.27 & 43.61 \\
 & \textbf{SFT+\framework{}} & \textbf{50.30} {\footnotesize (+33.62)} & \textbf{66.10} {\footnotesize (+4.18)} & \textbf{37.88} {\footnotesize (+4.65)} & \textbf{54.08} {\footnotesize (+14.34)} & \textbf{52.09} {\footnotesize (+14.20)} \\
\midrule

\multirow{5}{*}{Qwen3-4B-Thinking}
 & Zero-shot & 7.66 & 47.58 & 20.28 & 24.00 & 24.88 \\
 & SFT & 29.42 & 42.42 & 17.39 & 27.17 & 29.10 \\
 & SFT+CLIP & 24.62 & 39.14 & 19.53 & 21.55 & 26.21 \\
 & SFT+CompreCap & 32.20 & 30.04 & 17.40 & 29.18 & 27.21 \\
 & \textbf{SFT+\framework{}} & \textbf{33.02} {\footnotesize (+25.36)} & \textbf{50.82} {\footnotesize (+3.24)} & \textbf{22.29} {\footnotesize (+2.01)} & \textbf{30.23} {\footnotesize (+6.23)} & \textbf{34.09} {\footnotesize (+9.21)} \\
\midrule

\multirow{5}{*}{GLM-Flash}
 & Zero-shot & 10.44 & 41.41 & 20.19 & 24.32 & 24.09 \\
 & SFT & 32.78 & 43.17 & 23.72 & 34.51 & 33.54 \\
 & SFT+CLIP & 38.33 & 55.10 & 29.85 & 38.56 & 40.46 \\
 & SFT+CompreCap & 43.22 & 40.12 & 28.83 & 47.57 & 39.94 \\
 & \textbf{SFT+\framework{}} & \textbf{43.35} {\footnotesize (+32.91)} & \textbf{58.86} {\footnotesize (+17.45)} & \textbf{34.02} {\footnotesize (+13.83)} & \textbf{48.58} {\footnotesize (+24.26)} & \textbf{46.20} {\footnotesize (+22.11)} \\
\bottomrule
\end{tabular}
}

\caption{Fixed-judge \benchmark{} rescoring on InternVL3-2B under three evaluation judges. Parentheses denote gains over Zero-shot.}
\label{tab:scope_fixed_judges}

\end{table*}

%% file: latex/sec/A_Exp_Details.tex
\section{Additional Experimental and Training Details}
\label{app:exp_details}

\subsection{SFT and GRPO Data Construction}
\label{app:exp_details_data}

\subsubsection{SFT data construction}
For SFT, we randomly sample 2{,}000 images from the COCO 2017 training set~\citep{lin2014microsoft}. For each image, we use GPT-5~\citep{openai2025introducing} to generate five category-specific captions aligned with our control prompt categories (General, Attribute, Relation, Foreground, Background). The generated captions are manually filtered for quality and category alignment, yielding 10{,}000 image--caption pairs in total.

\subsubsection{GRPO data construction (CompreCap split and FG/BG labels)}
For GRPO, we use CompreCap, which provides dense human-annotated scene graphs (objects, segmentation masks, bound attributes, directional relations). From 560 instances, we randomly sample 280 for training and reserve the remaining 280 for evaluation (no overlap). To support foreground/background-focused control, we augment CompreCap objects with \textit{Foreground} vs.\ \textit{Background} labels using GPT-5; location descriptors derived from visual cues are consistently categorized as background. These foreground/background classifications are subsequently manually filtered.

\subsection{Implementation Details}
\label{app:exp_details_impl}

We implement GRPO using the Hugging Face \texttt{trl} library~\citep{Huggingface--trl} on 8 NVIDIA RTX6000 Ada 48GB GPUs. For GRPO, we use 8 generations per prompt, gradient accumulation steps of 8, KL coefficient $\beta=0.04$, and learning rate $2 \times 10^{-6}$. During training, we use a multi-prompt sampling strategy to balance optimization across the five prompt categories.
\input{latex/sec/tables/A_random_seed}

\subsection{Random Seed Stability}
\label{app:seed_stability}

To assess the stability of GRPO training, we repeated SFT+\framework{} with two additional random seeds for each backbone and report mean $\pm$ standard deviation over three runs. Overall, the results are stable across seeds on both \benchmark{} and the CompreCap evaluation: the standard deviation of the overall score remains small in all settings---below 0.5 for both backbones on both benchmarks.

We observe slightly larger variation in some harder fine-grained categories, particularly foreground/background-focused captioning, but these fluctuations are modest and do not affect the main conclusions of the paper. In particular, \framework{} shows consistent performance across runs for both Qwen2.5-VL-3B and InternVL3-2B, indicating that the reported improvements are not driven by a favorable single seed. Full per-seed results are provided in Table~\ref{tab:seed_analysis_combined}.

\subsection{\benchmark{} Robustness to Source Composition and Sampling}
\label{app:scope_robustness}

Because \benchmark{} contains images from multiple source datasets, we examine whether the observed gains are concentrated on images originating from CompreCap, which is also used for GRPO training. Of the 189 \benchmark{} images, 35 originate from CompreCap and the remaining 154 from COCO and DOCCI; none overlap with the SFT or GRPO training images. We report Overall scores separately for these two source groups. To quantify uncertainty due to the finite evaluation set, we additionally compute 95\% nonparametric image-bootstrap confidence intervals on the full \benchmark{} set using 20{,}000 repetitions.

\input{latex/sec/tables/A_SCoPE_Robustness}

As shown in Table~\ref{tab:scope_source_robustness}, SFT+\framework{} achieves the best Overall score on both source subsets for both backbones. In particular, its performance is similar between CompreCap-origin and non-CompreCap images, despite GRPO training using CompreCap annotations. On the full benchmark, its 95\% bootstrap interval is also separated from that of the strongest non-\framework{} baseline for each backbone. Together with the seed-stability results above, these results show that the observed gains are not concentrated in a particular \benchmark{} source subset or a favorable training run.

\subsection{Evaluation Variant on CompreCap}
\label{app:exp_details_eval}

In addition to \benchmark{}, we evaluate on the held-out 280 CompreCap instances by reweighting CompreCap components for different control emphases and scaling scores to 0--100 for interpretability. To address the limitations noted in the main text, we apply a stricter object validity threshold ($\tau=0.6$) and upgrade the scoring engine to \texttt{Qwen3-32B-Thinking}~\citep{yang2025qwen3} for more rigorous reasoning-based assessment.

When performing category-specific reweighting for this \emph{CompreCap-based evaluation variant}, we use a milder negative weight of $-0.25$ (instead of $-1.0$) on the off-target component (e.g., relations in attribute-focused scoring, and attributes in relation-focused scoring).

\input{latex/sec/tables/A_DOCCI_General}

\subsection{General Caption Evaluation}
\label{app:general_caption_eval}

To assess whether improved controllability comes at the expense of general caption quality, we additionally evaluate captions generated with the General prompt on 5{,}000 images from the DOCCI test split~\citep{onoe2024docci}. Using the human-written DOCCI captions as references, we report standard reference-based metrics---CIDEr~\citep{vedantam2015cidera}, METEOR~\citep{denkowski2014meteor}, and ROUGE-L~\citep{lin2004rouge}---together with CAPTURE~\citep{dong2024benchmarking} and the general-prompt CompreCap~\citep{lu2025benchmarking} score. All scores are reported on a $[0,1]$ scale.

As shown in Table~\ref{tab:general_caption}, SFT+\framework{} improves all reported metrics for Qwen2.5-VL-3B. For InternVL3-2B, METEOR and general-prompt CompreCap improve, CAPTURE remains essentially unchanged, and CIDEr and ROUGE-L decrease slightly. Overall, \framework{} largely maintains general-caption quality across the two backbones while consistently improving fine-grained factual alignment under the general prompt.

\subsection{Baselines (Expanded)}
\label{app:exp_details_baselines}

\begin{itemize}
    \item \textbf{Zero-shot:} the pretrained backbone with only task-specific prompts (no parameter updates).
    \item \textbf{SFT:} cross-entropy fine-tuning on the COCO-based category-specific caption dataset (Appendix~\ref{app:exp_details_data}).
    \item \textbf{SFT+CLIP:} after SFT, we add CLIP-based focus indicators to specify the target emphasis, then run GRPO using CLIP similarity as reward.
    \item \textbf{SFT+CompreCap:} GRPO using the original CompreCap scoring pipeline as the training reward, without the stricter object-matching threshold and reasoning-based verification for attribute and relation scoring used in \framework{}.
\end{itemize}

\subsection{LLM Judges Used in Training vs.\ Evaluation}
\label{app:exp_details_judges}

For the CompreCap-based evaluation variant, we use \texttt{Qwen3-32B-Thinking}~\citep{yang2025qwen3} as the scoring engine for more rigorous reasoning-based assessment.

%% file: latex/sec/tables/A_random_seed.tex
\begin{table*}[!t]

\centering

\resizebox{\textwidth}{!}{

\begin{tabular}{@{}llcccccc@{}}

\toprule

\textbf{Base Model} & \textbf{Seed} & \multicolumn{6}{c}{\textbf{\benchmark{}}} \\

\cmidrule(lr){3-8}

& & & \textbf{Attribute} & \textbf{Relation} & \textbf{Foreground} & \textbf{Background} & \textbf{Overall} \\

\midrule

\multirow{4}{*}{Qwen2.5-VL-3B} 
 & 0 & & 30.12 & 45.46 & 24.09 & 25.95 & 31.40 \\
 & 1 & & 30.31 & 46.12 & 23.54 & 26.76 & 31.68 \\
 & 2 & & 30.89 & 45.06 & 25.52 & 27.02 & 32.12 \\
 & \textbf{Mean $\pm$ std} & & \textbf{30.44 $\pm$ 0.40} & \textbf{45.55 $\pm$ 0.54} & \textbf{24.38 $\pm$ 1.02} & \textbf{26.58 $\pm$ 0.56} & \textbf{31.74 $\pm$ 0.36} \\

\midrule

\multirow{4}{*}{InternVL3-2B} 
 & 0 & & 36.18 & 54.39 & 25.18 & 31.68 & 36.86 \\
 & 1 & & 36.39 & 54.35 & 25.46 & 29.97 & 36.54 \\
 & 2 & & 37.13 & 53.88 & 25.23 & 32.47 & 37.18 \\
 & \textbf{Mean $\pm$ std} & & \textbf{36.57 $\pm$ 0.50} & \textbf{54.21 $\pm$ 0.28} & \textbf{25.29 $\pm$ 0.15} & \textbf{31.37 $\pm$ 1.28} & \textbf{36.86 $\pm$ 0.32} \\

\midrule
\midrule

& & \multicolumn{6}{c}{\textbf{CompreCap}} \\

\cmidrule(lr){3-8}

& & \textbf{General} & \textbf{Attribute} & \textbf{Relation} & \textbf{Foreground} & \textbf{Background} & \textbf{Overall} \\

\midrule

\multirow{4}{*}{Qwen2.5-VL-3B} 
 & 0 & 46.86 & 44.29 & 54.16 & 55.26 & 50.88 & 50.29 \\
 & 1 & 48.00 & 45.45 & 55.15 & 56.22 & 51.35 & 51.23 \\
 & 2 & 47.68 & 45.40 & 54.73 & 55.50 & 51.38 & 50.94 \\
 & \textbf{Mean $\pm$ std} & \textbf{47.51 $\pm$ 0.59} & \textbf{45.05 $\pm$ 0.65} & \textbf{54.68 $\pm$ 0.50} & \textbf{55.66 $\pm$ 0.50} & \textbf{51.20 $\pm$ 0.28} & \textbf{50.82 $\pm$ 0.48} \\

\midrule

\multirow{4}{*}{InternVL3-2B} 
 & 0 & 49.23 & 46.97 & 56.57 & 55.81 & 51.41 & 52.00 \\
 & 1 & 49.78 & 47.69 & 55.59 & 55.55 & 52.37 & 52.20 \\
 & 2 & 49.14 & 47.10 & 56.99 & 54.77 & 51.23 & 51.85 \\
 & \textbf{Mean $\pm$ std} & \textbf{49.38 $\pm$ 0.35} & \textbf{47.25 $\pm$ 0.38} & \textbf{56.38 $\pm$ 0.72} & \textbf{55.38 $\pm$ 0.54} & \textbf{51.67 $\pm$ 0.61} & \textbf{52.02 $\pm$ 0.18} \\

\bottomrule

\end{tabular}

}

\caption{Random seed analysis for Qwen2.5-VL-3B-Instruct and InternVL3-2B with SFT+\framework{}. We report results from three independent GRPO runs and the mean $\pm$ standard deviation for both \benchmark{} and CompreCap metrics.}

\label{tab:seed_analysis_combined}

\end{table*}

%% file: latex/sec/tables/A_SCoPE_Robustness.tex
\begin{table*}[!t]
\centering

\resizebox{\textwidth}{!}{
\begin{tabular}{@{}llccc@{}}
\toprule
\textbf{Base Model} & \textbf{Method} & \multicolumn{3}{c}{\textbf{\benchmark{} Overall}} \\
\cmidrule(lr){3-5}
& & \textbf{CompreCap-Origin} & \textbf{Non-CompreCap} & \textbf{All [95\% CI]} \\
\midrule

\multirow{5}{*}{Qwen2.5-VL-3B}
& Zero-shot
& 13.74 & 16.01 & 15.66 [14.21, 17.04] \\
& SFT
& 17.32 & 23.81 & 22.61 [20.61, 24.46] \\
& SFT+CLIP
& 27.90 & 26.03 & 26.40 [25.18, 27.62] \\
& SFT+CompreCap
& 25.81 & 25.86 & 25.85 [24.65, 27.06] \\
& \textbf{SFT+\framework{}}
& \textbf{31.74} & \textbf{31.75} & \textbf{31.74 [30.55, 32.88]} \\
\midrule

\multirow{5}{*}{InternVL3-2B}
& Zero-shot
& 24.31 & 25.93 & 25.63 [24.60, 26.63] \\
& SFT
& 32.05 & 31.01 & 31.21 [29.97, 32.42] \\
& SFT+CLIP
& 27.94 & 28.49 & 28.39 [27.18, 29.57] \\
& SFT+CompreCap
& 30.44 & 28.99 & 29.26 [28.18, 30.31] \\
& \textbf{SFT+\framework{}}
& \textbf{37.90} & \textbf{36.62} & \textbf{36.86 [35.79, 37.91]} \\
\bottomrule
\end{tabular}
}

\caption{Source-stratified evaluation on \benchmark{}. We report Overall scores separately for CompreCap-origin images and images originating from the other source datasets, together with 95\% nonparametric image-bootstrap confidence intervals on the full benchmark using 20{,}000 repetitions.}
\label{tab:scope_source_robustness}
\end{table*}

%% file: latex/sec/tables/A_DOCCI_General.tex
\begin{table*}[!t]
\centering

\resizebox{\textwidth}{!}{
\begin{tabular}{@{}llccccc@{}}
\toprule
\textbf{Base Model} & \textbf{Method} & \multicolumn{5}{c}{\textbf{General Captioning on DOCCI}} \\
\cmidrule(lr){3-7}
& & \textbf{CIDEr} & \textbf{METEOR} & \textbf{ROUGE-L} & \textbf{CAPTURE} & \textbf{CompreCap (General)} \\
\midrule

\multirow{2}{*}{Qwen2.5-VL-3B}
& Zero-shot
& 0.0040 & 0.1273 & 0.1694 & 0.5627 & 0.4162 \\
& \textbf{SFT+\framework{}}
& 0.0879 & 0.1319 & 0.2062 & 0.5981 & 0.4751 \\
\midrule

\multirow{2}{*}{InternVL3-2B}
& Zero-shot
& 0.1110 & 0.1415 & 0.2196 & 0.5989 & 0.3999 \\
& \textbf{SFT+\framework{}}
& 0.0908 & 0.1490 & 0.2118 & 0.5990 & 0.4938 \\
\bottomrule
\end{tabular}
}

\caption{General-caption evaluation on 5{,}000 images from the DOCCI test split. We report standard reference-based metrics together with CAPTURE and general-prompt CompreCap. All scores are scaled to $[0,1]$.}
\label{tab:general_caption}
\end{table*}

%% file: latex/sec/A_more_ablations.tex
\section{Additional Controllability Analyses}
\label{app:control_ablations}

\input{latex/sec/tables/A_prompt_eng}
\subsection{Prompt-Engineering Ablation on \benchmark{}}
\label{app:prompt_engineering}

We additionally evaluate all methods using stronger inference prompts that explicitly specify both desired and off-scope content. For example, the background-focused prompt instructs the model to describe only background elements and to exclude foreground subjects, body parts, actions, and interactions; analogous constraints are used for attribute-, relation-, and foreground-focused captioning. This ablation changes only the inference prompt and leaves the underlying model parameters unchanged.

As shown in Table~\ref{tab:prompt_eng_main}, SFT+\framework{} achieves the highest Overall score on both backbones and performs best across all four control categories. These results complement the main evaluation by showing that prompt-conditioned training provides gains even when the desired behavior is specified with explicit inference-time instructions.

\input{latex/sec/tables/A_weight_abl}
\subsection{Ablating Prompt-Conditioned and Contrastive Reward Weighting}
\label{app:weighting_ablation}

We further ablate the reward-weighting design introduced in Sec.~\ref{sec:method} and evaluate on \benchmark{} (Sec.~\ref{sec:benchmark}). To isolate the effect of weighting, for each backbone we keep the training and evaluation setup fixed and vary only the reward aggregation. We compare \framework{} against two alternatives: \textbf{Fixed Reward Weights}, which applies the same general-caption mixture to every prompt category, and \textbf{Simply Up-weighted}, which keeps category-specific target emphasis but removes negative weights on non-target components. Results are shown in Table~\ref{tab:weighting_ablation}.

Across both Qwen2.5-VL-3B and InternVL3-2B, both alternatives underperform \framework{} in Overall score and show a less balanced category profile. The ablated variants can obtain higher Foreground scores than \framework{}, but they substantially lag on Attribute, Relation, and Background. This suggests that using either a prompt-agnostic mixture or target-only positive weighting tends to favor salient foreground or generic scene content across prompt types. With Fixed Reward Weights, the same aggregation is used for every prompt category; with Simply Up-weighted, target components receive more weight but off-scope components are not explicitly suppressed. Under \benchmark{}'s Include/Avoid evaluation, these design choices lead to weaker cross-category controllability.

By contrast, \framework{} achieves the strongest Overall score on both backbones and the best Attribute, Relation, and Background scores, supporting the role of prompt-conditioned signed weights in jointly encouraging on-scope content and suppressing off-scope content.

\input{latex/sec/tables/main_table_commercial}

\input{latex/sec/figures/penalty}
\subsection{Penalty Magnitude Sensitivity}
\label{app:penalty_sensitivity}

Figure~\ref{fig:penalty} shows how the magnitude of the negative reward penalty shapes behavior. As the penalty increases, behavior can shift sharply: very large penalties may produce refusal-like outputs, while removing the penalty ($0.0$) increases scope drift and invites off-scope background details. This suggests using a bounded, calibrated penalty rather than extreme values; accordingly, we adopt a moderate default that reduces off-scope content without encouraging refusals.

%% file: latex/sec/tables/A_prompt_eng.tex
\begin{table*}[!t]
\centering

\resizebox{\textwidth}{!}{
\begin{tabular}{@{}llccccc@{}}
\toprule
\textbf{Base Model} & \textbf{Method} & \multicolumn{5}{c}{\textbf{\benchmark{}}} \\
\cmidrule(lr){3-7}
& & \textbf{Attribute} & \textbf{Relation} & \textbf{Foreground} & \textbf{Background} & \textbf{Overall} \\
\midrule

\multirow{5}{*}{Qwen2.5-VL-3B}
& Zero-shot & 5.43 & 16.08 & 13.66 & 16.04 & 12.80 \\
& SFT & 20.97 & 33.72 & 19.94 & 18.25 & 23.22 \\
& SFT+CLIP & 22.15 & 40.31 & 25.35 & 21.69 & 27.37 \\
& SFT+CompreCap & 27.68 & 31.33 & 23.63 & 26.59 & 27.31 \\
& \textbf{SFT+\framework{}} & \textbf{28.36} {\footnotesize (+22.93)} & \textbf{45.66} {\footnotesize (+29.58)} & \textbf{29.10} {\footnotesize (+15.44)} & \textbf{27.12} {\footnotesize (+11.08)} & \textbf{32.56} {\footnotesize (+19.76)} \\
\midrule

\multirow{5}{*}{InternVL3-2B}
& Zero-shot & 9.02 & 28.32 & 21.64 & 22.87 & 20.46 \\
& SFT & 28.05 & 44.93 & 23.90 & 27.14 & 31.01 \\
& SFT+CLIP & 19.64 & 41.36 & 24.74 & 21.85 & 26.90 \\
& SFT+CompreCap & 29.78 & 32.83 & 22.15 & 29.22 & 28.50 \\
& \textbf{SFT+\framework{}} & \textbf{32.26} {\footnotesize (+23.24)} & \textbf{50.79} {\footnotesize (+22.47)} & \textbf{29.50} {\footnotesize (+7.86)} & \textbf{29.95} {\footnotesize (+7.08)} & \textbf{35.62} {\footnotesize (+15.16)} \\
\bottomrule
\end{tabular}
}

\caption{\benchmark{} results under engineered prompts for Qwen2.5-VL-3B-Instruct and InternVL3-2B. Parentheses denote gains over Zero-shot.}
\label{tab:prompt_eng_main}
\end{table*}

%% file: latex/sec/tables/A_weight_abl.tex
\begin{table*}[!t]
\centering

\resizebox{\textwidth}{!}{
\begin{tabular}{@{}llccccc@{}}
\toprule
\textbf{Base Model} & \textbf{Method} & \multicolumn{5}{c}{\textbf{\benchmark{}}} \\
\cmidrule(lr){3-7}
& & \textbf{Attribute} & \textbf{Relation} & \textbf{Foreground} & \textbf{Background} & \textbf{Overall} \\
\midrule

\multirow{3}{*}{Qwen2.5-VL-3B}
& Fixed Reward Weights & 9.40 & 36.77 & \textbf{28.38} & 18.13 & 23.17 \\
& Simply Up-weighted & 9.42 & 33.92 & 27.07 & 18.97 & 22.35 \\
& \textbf{\framework{}} & \textbf{30.44} & \textbf{45.55} & 24.38 & \textbf{26.58} & \textbf{31.74} \\
\midrule

\multirow{3}{*}{InternVL3-2B}
& Fixed Reward Weights & 12.68 & 40.96 & \textbf{33.66} & 27.09 & 28.60 \\
& Simply Up-weighted & 11.58 & 40.07 & 31.49 & 26.20 & 27.33 \\
& \textbf{\framework{}} & \textbf{36.57} & \textbf{54.21} & 25.29 & \textbf{31.37} & \textbf{36.86} \\
\bottomrule
\end{tabular}
}

\caption{Ablation of reward-weighting design on \benchmark{}.}
\label{tab:weighting_ablation}
\end{table*}

%% file: latex/sec/tables/main_table_commercial.tex
\begin{table*}[!t]

\centering

\resizebox{\textwidth}{!}{

\begin{tabular}{@{}llcccccc@{}}

\toprule

\textbf{Base Model} & \textbf{Method} & \multicolumn{6}{c}{\textbf{\benchmark{}}} \\

\cmidrule(lr){3-8}

& & & \textbf{Attribute} & \textbf{Relation} & \textbf{Foreground} & \textbf{Background} & \textbf{Overall} \\

\midrule

\multirow{3}{*}{GPT-5} 

 & nano & & 17.93 & 63.84 & 39.21 & 43.82 & 41.20 \\

 & mini & & 25.34 & 70.43 & 45.54 & 50.57 & 47.97 \\

 & - & & \textbf{27.55} & \textbf{71.60} & \textbf{45.59} & \textbf{54.21} & \textbf{49.74} \\

\midrule

\multirow{3}{*}{Gemini 2.5} 

 & Flash Lite & & 18.52 & 60.15 & 41.96 & 36.73 & 39.34 \\

 & Flash & & 25.02 & 66.89 & 53.75 & 48.95 & 48.65 \\

 & Pro & & \textbf{30.25} & \textbf{70.64} & \textbf{60.40} & \textbf{62.15} & \textbf{55.86} \\

\midrule

\midrule

& & \multicolumn{6}{c}{\textbf{CompreCap}} \\

\cmidrule(lr){3-8}

& & \textbf{General} & \textbf{Attribute} & \textbf{Relation} & \textbf{Foreground} & \textbf{Background} & \textbf{Overall} \\

\midrule

\multirow{3}{*}{GPT-5} 

 & nano & 45.31 & \textbf{40.49} & 54.48 & 50.88 & 49.45 & 48.12 \\

 & mini & \textbf{47.96} & 39.89 & \textbf{59.27} & \textbf{51.40} & 51.81 & \textbf{50.07} \\

 & - & 45.68 & 40.15 & 56.95 & 49.45 & \textbf{53.27} & 49.10 \\

\midrule

\multirow{3}{*}{Gemini 2.5} 

 & Flash Lite & 45.32 & 39.23 & 52.91 & 51.33 & 47.11 & 47.18 \\

 & Flash & \textbf{52.14} & \textbf{43.43} & \textbf{58.30} & 51.35 & \textbf{51.91} & \textbf{51.42} \\

 & Pro & 51.11 & 42.13 & 56.65 & \textbf{53.75} & 51.33 & 50.99 \\

\bottomrule

\end{tabular}

}

\caption{Performance of commercial LVLMs on \benchmark{} and CompreCap evaluation protocols. Best results within each model family are shown in \textbf{bold}.}

\label{tab:main_table_commercial}

\end{table*}

%% file: latex/sec/figures/penalty.tex
\begin{figure}[!t]
\vspace{-2mm}
  \centering
  \includegraphics[width=\columnwidth]{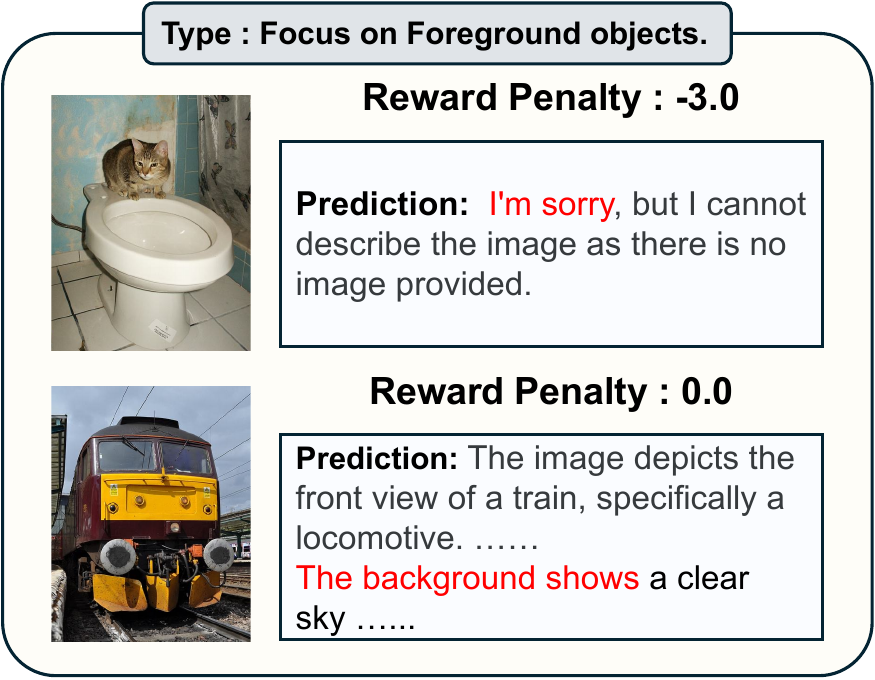}
   \caption{Penalty magnitude ablation for foreground-focused rewards. Large negative penalties (e.g., $-3.0$) can trigger refusal-like captions, while no penalty ($0.0$) causes scope drift by adding extra background details.}
  \label{fig:penalty}
  \vspace{-10pt}
\end{figure}

%% file: latex/sec/A_Commercial.tex
\section{\benchmark{} and CompreCap Benchmark for Commercial Models}
\label{app:commercial_bench}

Table~\ref{tab:main_table_commercial} reports results for representative commercial LVLMs on both evaluation protocols used in the main paper: \benchmark{} for contrastive controllability via Include/Avoid constraints and CompreCap for component-wise factual alignment.

\paragraph{\benchmark{}: Controllability under Include/Avoid constraints.}
Across model families, larger variants achieve higher controllability scores. For GPT-5, performance increases from nano (Overall 41.20) to mini (47.97) and peaks at the largest setting (Overall 49.74), with particularly strong Relation control (up to 71.60). Gemini~2.5 exhibits a similar scaling trend and attains the best overall \benchmark{} performance among the tested commercial models: Flash Lite (39.34) $<$ Flash (48.65) $<$ Pro (55.86). Gemini~2.5 Pro also leads category-wise, especially on Foreground (60.40) and Background (62.15), indicating stronger spatial emphasis and out-of-scope suppression in the contrastive setting.

\paragraph{CompreCap: Fine-grained factual alignment.}
On CompreCap, scores exhibit lower variance across model variants than on \benchmark{}, suggesting that component-wise factual alignment is less sensitive to model size than strict controllability. For GPT-5, the best overall CompreCap score is achieved by mini (50.07), while nano performs best on Attribute (40.49) and the largest setting leads on Background (53.27). For Gemini~2.5, Flash achieves the best overall CompreCap score (51.42) and leads on General (52.14), Attribute (43.43), Relation (58.30), and Background (51.91), while Pro performs best on Foreground (53.75).

\paragraph{Connection to main findings.}
These results contextualize the gains of \framework{}: \benchmark{} imposes stricter, contrastive control requirements than component-wise evaluation alone, and models that perform well on CompreCap do not always dominate on \benchmark{}. This gap motivates our prompt-conditioned reward weighting (including negative weights) and the emphasis on suppressing off-scope content, which is explicitly measured by \benchmark{} but only indirectly reflected in CompreCap.

%% file: latex/sec/A_Prompt.tex
\section{Prompt List}
\label{app:prompts}

This appendix lists the prompts used in \framework{} and for \benchmark{} evaluation.
The Attribute Evaluation Prompt and Relation Evaluation Prompt are used during training
to compute component-level scene rewards. The \benchmark{} inference prompts are used
to generate captions under each semantic control type. The Coverage-and-Adherence
Evaluation Prompt and Faithfulness Evaluation Prompt are used by the LLM judge during
\benchmark{} evaluation.

\subsection{Training Reward Judge Prompts}
\label{app:training_reward_prompts}

\begin{tcolorbox}[title=Attribute Evaluation Prompt, colback=gray!5, colframe=gray!75, breakable]
\small\ttfamily
\textbf{System Prompt:}\\
You are a strict evaluation judge. You will be given a `Sentence' (prediction) and a `Phrase' (ground truth description). Your task is to rate how accurately and completely the Sentence covers the detailed attributes (e.g., color, shape, material, design) described in the Phrase. \\[0.5em]
Rate on a scale from 0 to 5:\\
0: Irrelevant or contradictory.\\
1: Mentions the correct object but misses all specific attributes.\\
2: Captures the object and very minor details.\\
3: Captures the object and some key attributes.\\
4: Captures the object and most attributes.\\
5: Perfectly captures the object and all attributes described in the Phrase.\\[0.5em]
Explain the reason and then output the final score within \textless score\textgreater\textless/score\textgreater.\\[1em]
\textbf{Input Template:}\\
Sentence: \{sentence\}. Phrase: \{phrase\}.
\end{tcolorbox}

\begin{tcolorbox}[title=Relation Evaluation Prompt, colback=gray!5, colframe=gray!75, breakable]
\small\ttfamily
\textbf{System Prompt:}\\
You are a strict evaluation judge. You will be given a `Sentence' (prediction) and a `Phrase' (ground truth spatial relationship). Your task is to rate how accurately the Sentence describes the spatial relationship or interaction defined in the Phrase. \\[0.5em]
Rate on a scale from 0 to 5:\\
0: Relationship is missing, incorrect, or unrelated.\\
1: Mentions the objects but the relationship is wrong.\\
2: Mentions the objects and only weakly implies the relationship.\\
3: Captures the relationship vaguely or partially.\\
4: Correctly describes the relationship, but misses minor nuances or lacks full precision.\\
5: Perfectly captures the spatial relationship or interaction.\\[0.5em]
Explain the reason and then output the final score within \textless score\textgreater\textless/score\textgreater.\\[1em]
\textbf{Input Template:}\\
Sentence: \{sentence\}. Phrase: \{phrase\}.
\end{tcolorbox}

\subsection{\benchmark{} Inference Prompts}
\label{app:scope_inference_prompts}

\begin{tcolorbox}[
  title=\benchmark{} Inference Prompts,
  colback=gray!5, colframe=gray!75, breakable
]
\small
\textbf{Attribute:} ``Describe the image in detail, focusing on the attributes and characteristics of the objects.''

\medskip
\textbf{Relation:} ``Describe the image in detail, focusing on the spatial relationships between objects.''

\medskip
\textbf{Foreground:} ``Describe the image in detail, focusing on the foreground subject.''

\medskip
\textbf{Background:} ``Describe the image in detail, focusing on the background elements and surrounding environment.''
\end{tcolorbox}

\subsection{\benchmark{} LLM-Judge Prompts}
\label{app:scope_judge_prompts}

\begin{tcolorbox}[title=Coverage-and-Adherence Evaluation Prompt, colback=gray!5, colframe=gray!75, breakable]
\small\ttfamily
\textbf{Prompt:}\\
Does this caption mention the following?\\[0.5em]

Caption: "\{caption\}"\\
Mention: "\{fact\}"\\[0.5em]

Think carefully and answer Yes or No.
\end{tcolorbox}

\begin{tcolorbox}[title=Faithfulness Evaluation Prompt, colback=gray!5, colframe=gray!75, breakable]
\small\ttfamily
\textbf{Prompt:}\\
Does the caption describe something DIFFERENTLY than the fact?\\
(e.g., if fact says ``red ball'' but caption says ``blue ball'', that's a contradiction)\\[0.5em]

Caption: ``\{caption\}''\\
Fact: ``\{fact\}''\\[0.5em]

Think carefully and answer Yes or No.
\end{tcolorbox}

%% file: latex/sec/A_Examples.tex
\section{Qualitative Examples of \framework{}}
\label{app:focus_examples}

Figure~\ref{fig:focus_example1} shows a qualitative comparison between zero-shot and \framework{} generations across different control categories.
\input{latex/sec/figures/focus_example1}

%% file: latex/sec/figures/focus_example1.tex
\begin{figure*}[ht]
  \centering
  \includegraphics[width=\textwidth]{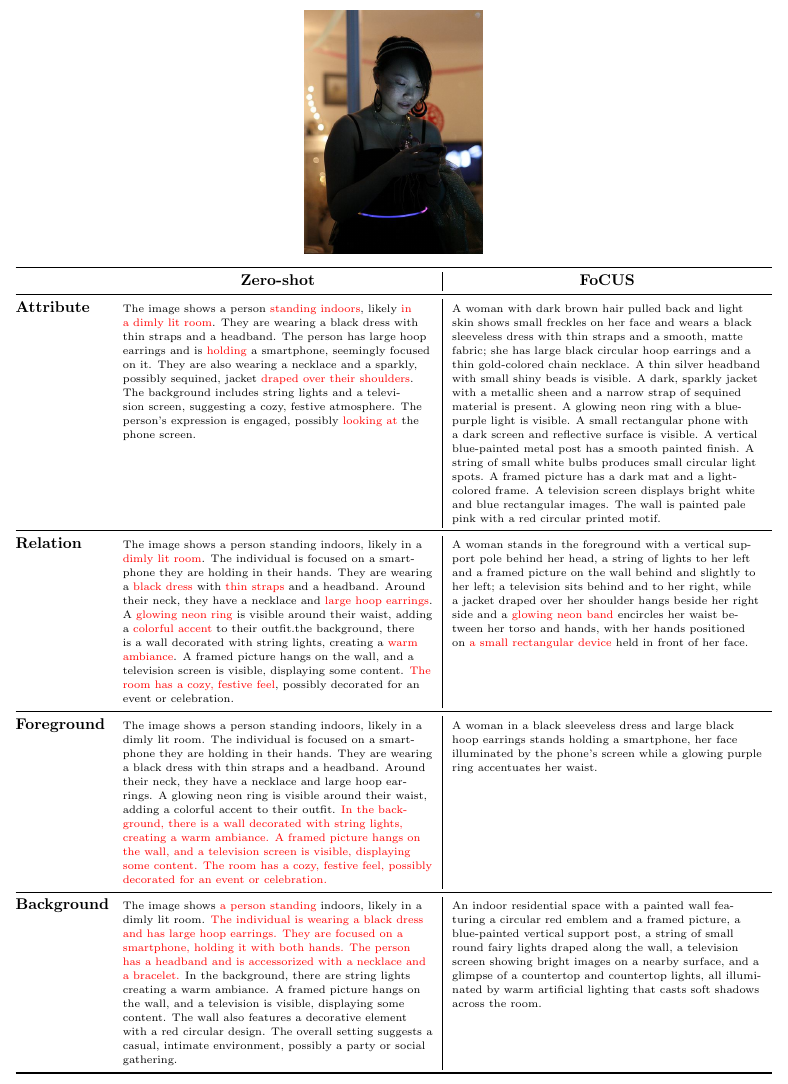}
  \caption{Zero-shot vs.\ \framework{} qualitative example. Rows correspond to control categories (Attribute, Relation, Foreground, Background). Red highlights out-of-category content that should be avoided under the given control signal. Compared to zero-shot, \framework{} better concentrates on the requested semantics while suppressing off-scope details.}
\label{fig:focus_example1}
\end{figure*}